\documentclass[10pt,twocolumn,letterpaper]{article}
\pdfoutput=1
\usepackage{cvpr}
\usepackage{times}
\usepackage{epsfig}
\usepackage{graphicx}
\usepackage{amsmath}
\usepackage{amssymb}
\usepackage{cite}
\usepackage{caption}
\usepackage{subcaption}
\graphicspath{ {figs/} }
\usepackage{booktabs}
\usepackage{multicol}
\usepackage{enumitem}
\usepackage{arydshln}
\usepackage{colortbl}
\usepackage{arydshln,xcolor,array}
\usepackage{bm}

\newlength\lengtha 
\usepackage[pagebackref=true,breaklinks=true,letterpaper=true,colorlinks,bookmarks=false]{hyperref}

\DeclareMathOperator*{\KC}{KC}
\DeclareMathOperator*{\K}{K}
\def\x{\mathbf{x}}

\cvprfinalcopy 

\ifcvprfinal\fi
\begin{document}

\title{Mining Point Cloud Local Structures by Kernel Correlation and Graph Pooling}

\author{
 {Yiru Shen\thanks{The authors contributed equally. This work is supported by MERL.} \ \footnotemark[2]{}}\\
  {\tt\small yirus@g.clemson.edu}
 \and
 {Chen Feng\footnotemark[1]{} \ \footnotemark[3]{}}\\
  {\tt\small cfeng@merl.com}
 \and
 {Yaoqing Yang\footnotemark[4]{}}\\
  {\tt\small yyaoqing@andrew.cmu.edu}
 \and
 {Dong Tian\footnotemark[3]{}}\\
  {\tt\small tian@merl.com}
 \and
 \normalsize{\textsuperscript{$\dagger$}Clemson University \quad \textsuperscript{$\ddagger$}Mitsubishi Electric Research Laboratories (MERL) \quad \textsuperscript{$\mathsection$}Carnegie Mellon University}
}


\maketitle

\thispagestyle{empty}

\begin{abstract}
   Unlike on images, semantic learning on 3D point clouds using a deep network is challenging due to the naturally unordered data structure. Among existing works, PointNet has achieved promising results by directly learning on point sets. However, it does not take full advantage of a point's local neighborhood that contains fine-grained structural information which turns out to be helpful towards better semantic learning. In this regard, we present two new operations to improve PointNet with a more efficient exploitation of local structures. The first one focuses on local 3D \emph{geometric} structures. In analogy to a convolution kernel for images, we define a point-set kernel as a set of learnable 3D points that jointly respond to a set of neighboring data points according to their geometric affinities measured by kernel correlation, adapted from a similar technique for point cloud registration. The second one exploits local high-dimensional \emph{feature} structures by recursive feature aggregation on a nearest-neighbor-graph computed from 3D positions. Experiments show that our network can efficiently capture local information and robustly achieve better performances on major datasets. Our code is available at \url{http://www.merl.com/research/license#KCNet}
\end{abstract}

\section{Introduction} \label{introduction}

\begin{figure}[!h]
    \begin{tabular}{@{}c@{}c@{}c@{}c@{}c@{}}
    	\includegraphics[width=0.2\columnwidth,height=1.3cm,keepaspectratio]{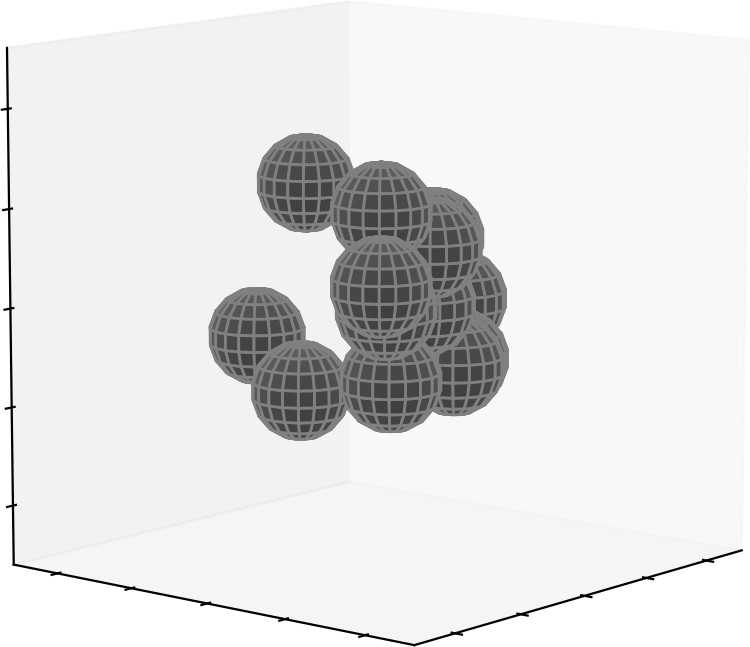} &
    	\includegraphics[width=0.2\columnwidth,height=1.3cm,keepaspectratio]{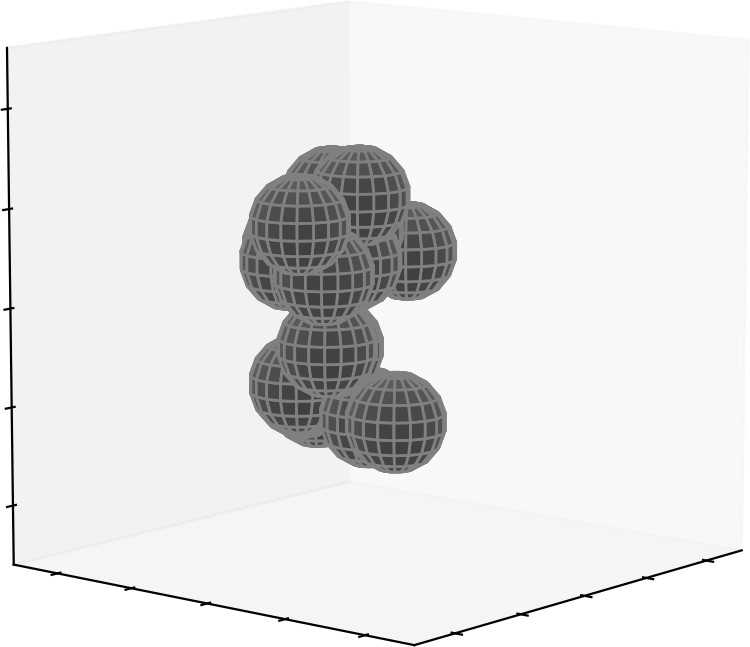} &
    	\includegraphics[width=0.2\columnwidth,height=1.3cm,keepaspectratio]{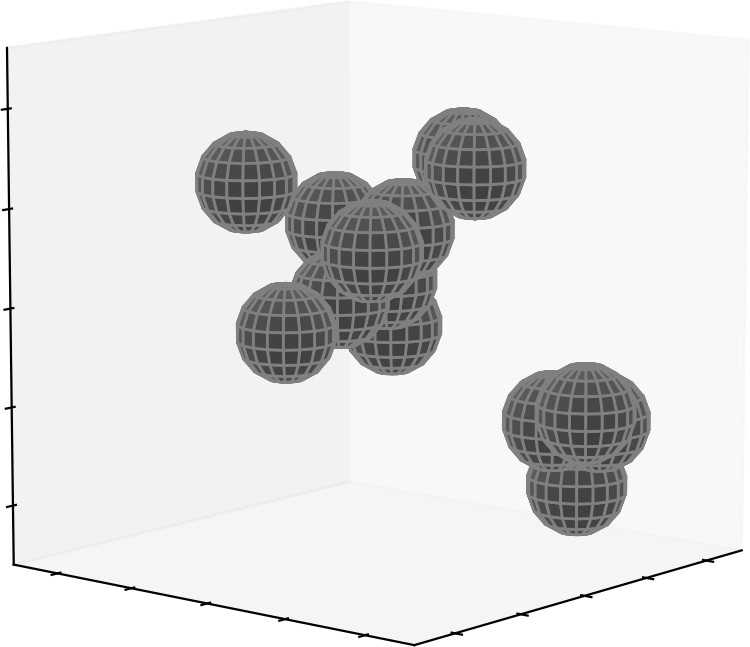} &
    	\includegraphics[width=0.2\columnwidth,height=1.3cm,keepaspectratio]{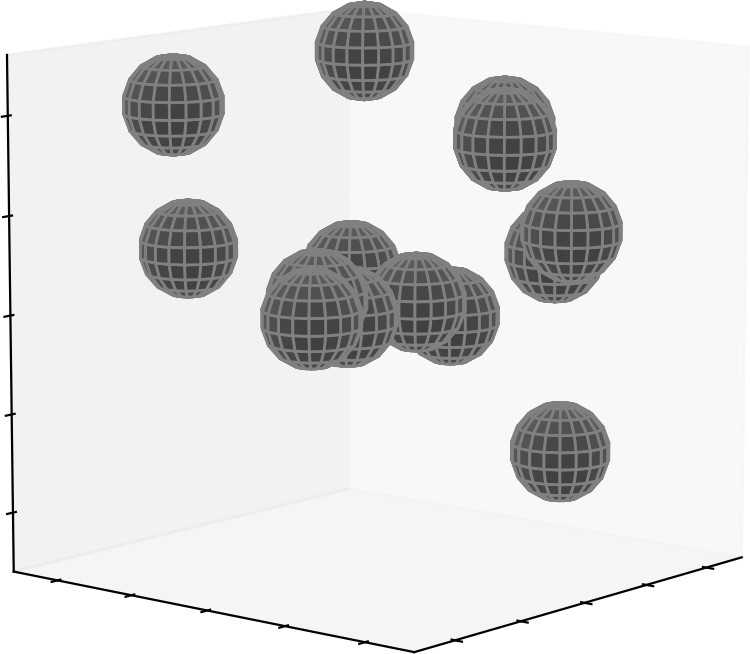} &
    	\includegraphics[width=0.2\columnwidth,height=1.3cm,keepaspectratio]{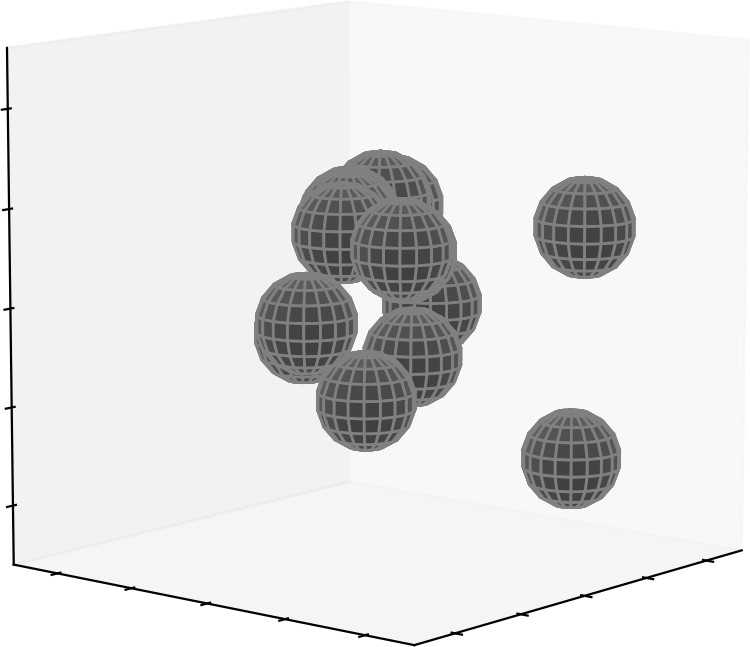} \\    	
   	
   		\includegraphics[width=0.2\columnwidth,height=1.3cm,keepaspectratio]{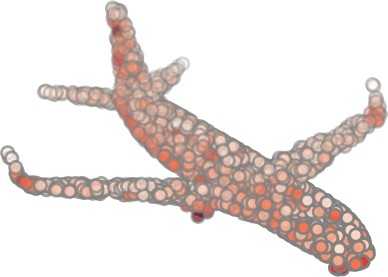} &
   		\includegraphics[width=0.2\columnwidth,height=1.3cm,keepaspectratio]{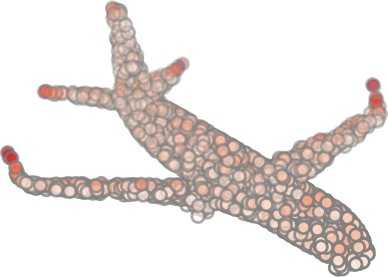} &
   		\includegraphics[width=0.2\columnwidth,height=1.3cm,keepaspectratio]{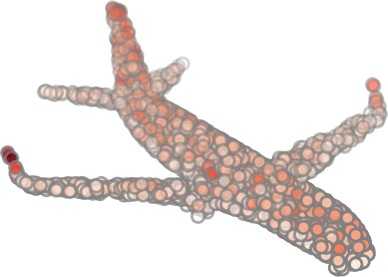} &
   		\includegraphics[width=0.2\columnwidth,height=1.3cm,keepaspectratio]{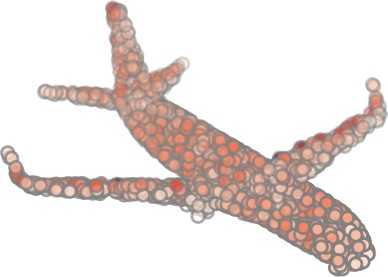} &
   		\includegraphics[width=0.2\columnwidth,height=1.3cm,keepaspectratio]{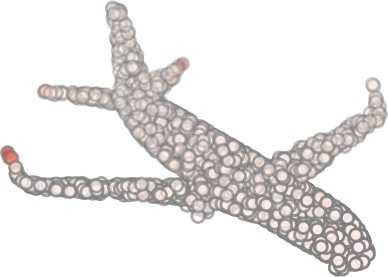}  \\ 
   		
    	\includegraphics[width=0.2\columnwidth,height=1.3cm,keepaspectratio]{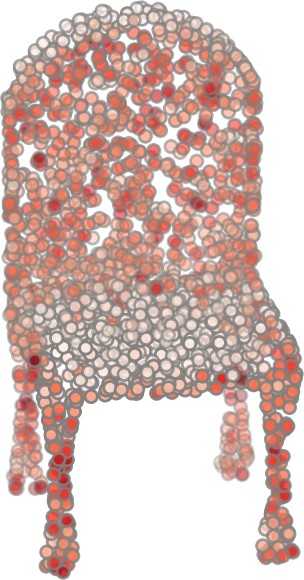} &
    	\includegraphics[width=0.2\columnwidth,height=1.3cm,keepaspectratio]{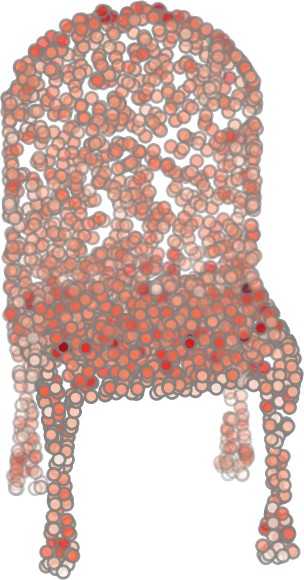} &
    	\includegraphics[width=0.2\columnwidth,height=1.3cm,keepaspectratio]{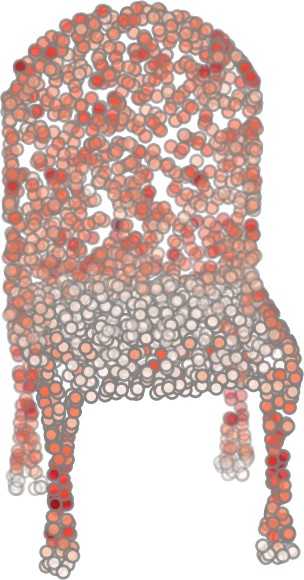} &
    	\includegraphics[width=0.2\columnwidth,height=1.3cm,keepaspectratio]{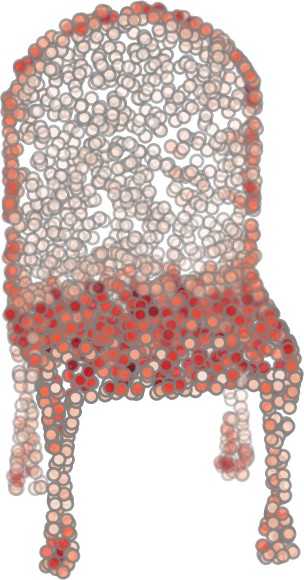} &
    	\includegraphics[width=0.2\columnwidth,height=1.3cm,keepaspectratio]{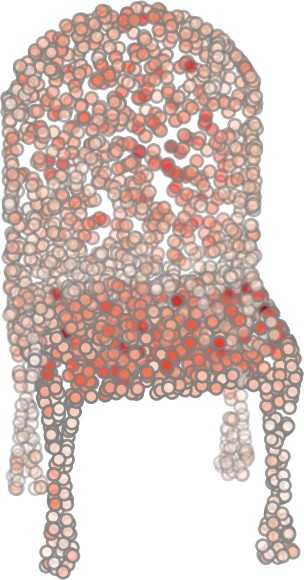}  \\ 
    	
    	\includegraphics[width=0.2\columnwidth,height=1.3cm,keepaspectratio]{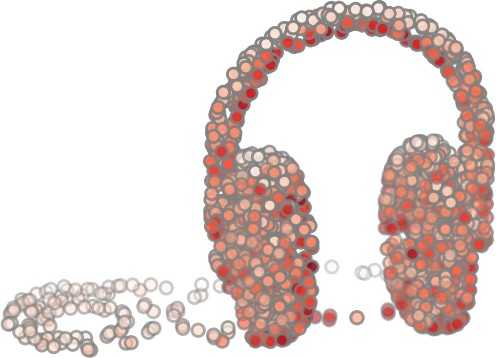} &
    	\includegraphics[width=0.2\columnwidth,height=1.3cm,keepaspectratio]{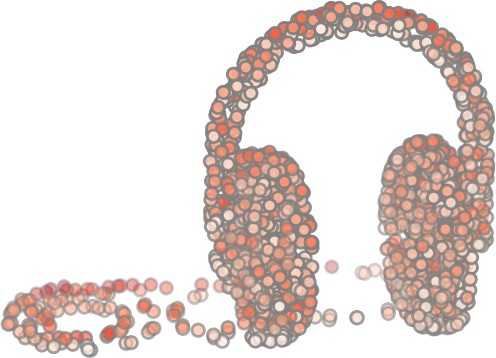} &
    	\includegraphics[width=0.2\columnwidth,height=1.3cm,keepaspectratio]{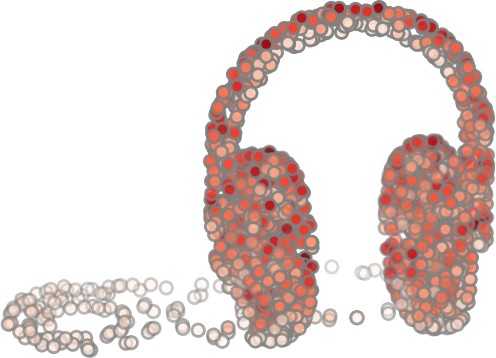} &
    	\includegraphics[width=0.2\columnwidth,height=1.3cm,keepaspectratio]{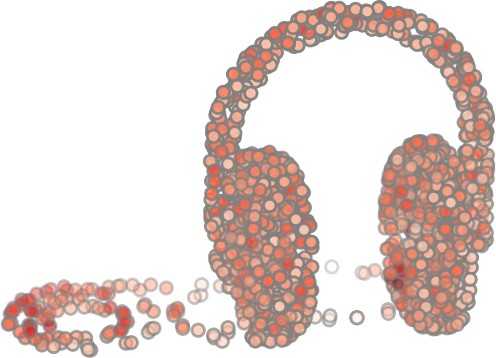} &
    	\includegraphics[width=0.2\columnwidth,height=1.3cm,keepaspectratio]{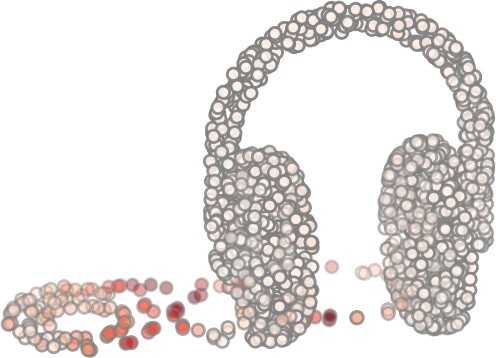}  \\ 
    	
    	\includegraphics[width=0.2\columnwidth,height=1.3cm,keepaspectratio]{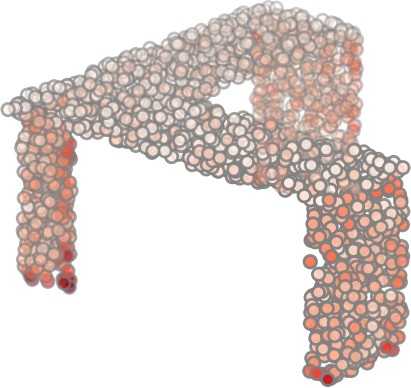} &
    	\includegraphics[width=0.2\columnwidth,height=1.3cm,keepaspectratio]{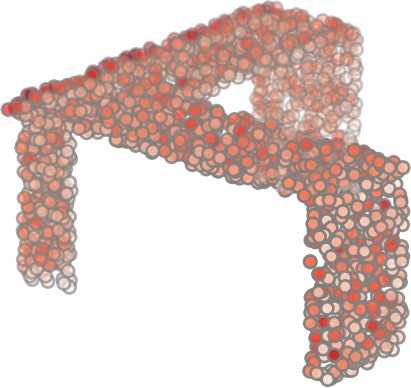} &
    	\includegraphics[width=0.2\columnwidth,height=1.3cm,keepaspectratio]{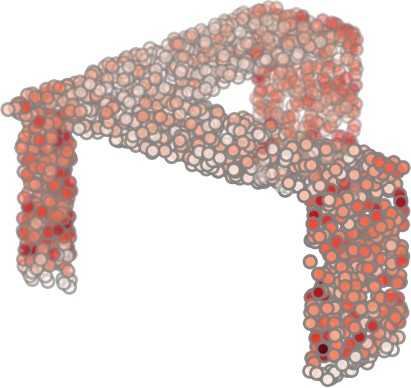} &
    	\includegraphics[width=0.2\columnwidth,height=1.3cm,keepaspectratio]{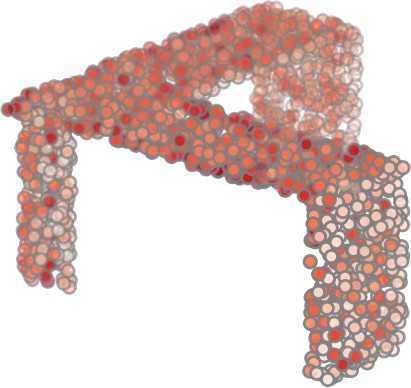} &
    	\includegraphics[width=0.2\columnwidth,height=1.3cm,keepaspectratio]{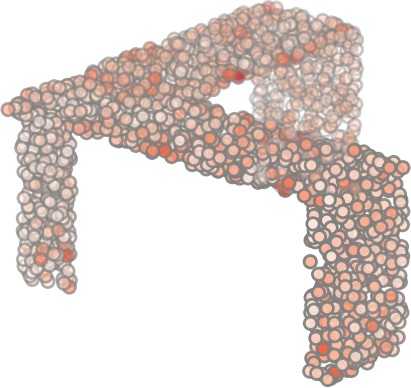}  \\ 
    \end{tabular} \vspace{-4mm}
    \caption{\textbf{Visualization of learned kernel correlations}. To represent complex local geometric structures around a point $p$, we propose kernel correlation as an affinity measure between two point sets: $p$'s neighboring points and kernel points. This figure shows kernel point positions and width as sphere centers and radius (top row), and the corresponding filter responses (other rows) of 5 kernels over 4 objects. Colors indicate affinities normalized in each object (red: strongest, white: weakest). Note the various structures (plane, edge, corner, concave and convex surfaces) captured by different kernels. Best viewed in color.}
    \label{fig:visualize_KC}
\end{figure}

As 3D data become ubiquitous with the rapid development of various 3D sensors, semantic understanding and analysis of such kind of data using deep networks is gaining attentions~\cite{qi2017pointnet,Wang2017OCNN,klokov2017escape,riegler2017octnet,bronstein2017geometric},
due to its wide applications in robotics, autonomous driving, reverse engineering, and civil infrastructure monitoring.
In particular, as one of the most primitive 3D data format and often the raw 3D sensor output, 3D point clouds cannot be trivially consumed by deep networks in the same way as 2D images by convolutional networks.
This is mainly caused by the irregular organization of points, a fundamental challenge inherent in this raw data format: compared with a row-column indexed image, a point cloud is a set of point coordinates (possibly with attributes like intensities and surface normals) without obvious orderings between points, except for point clouds computed from depth images.

\begin{figure*}
 \begin{subfigure}[b]{0.5\textwidth}
   \centering
   \includegraphics[width=\textwidth]{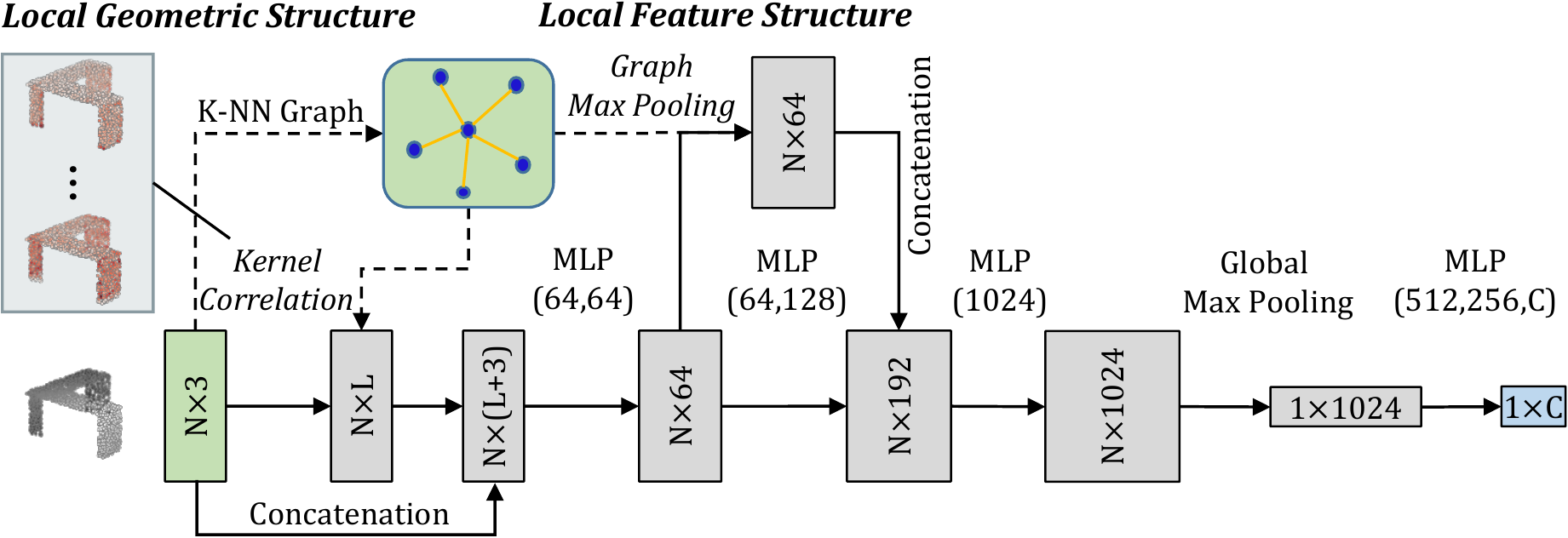}
   \caption{Classification network.}    
   \label{fig:arch-classification}      
  \end{subfigure}
  \hfill
  \begin{subfigure}[b]{0.46\textwidth}
  \centering
        \includegraphics[width=\textwidth]{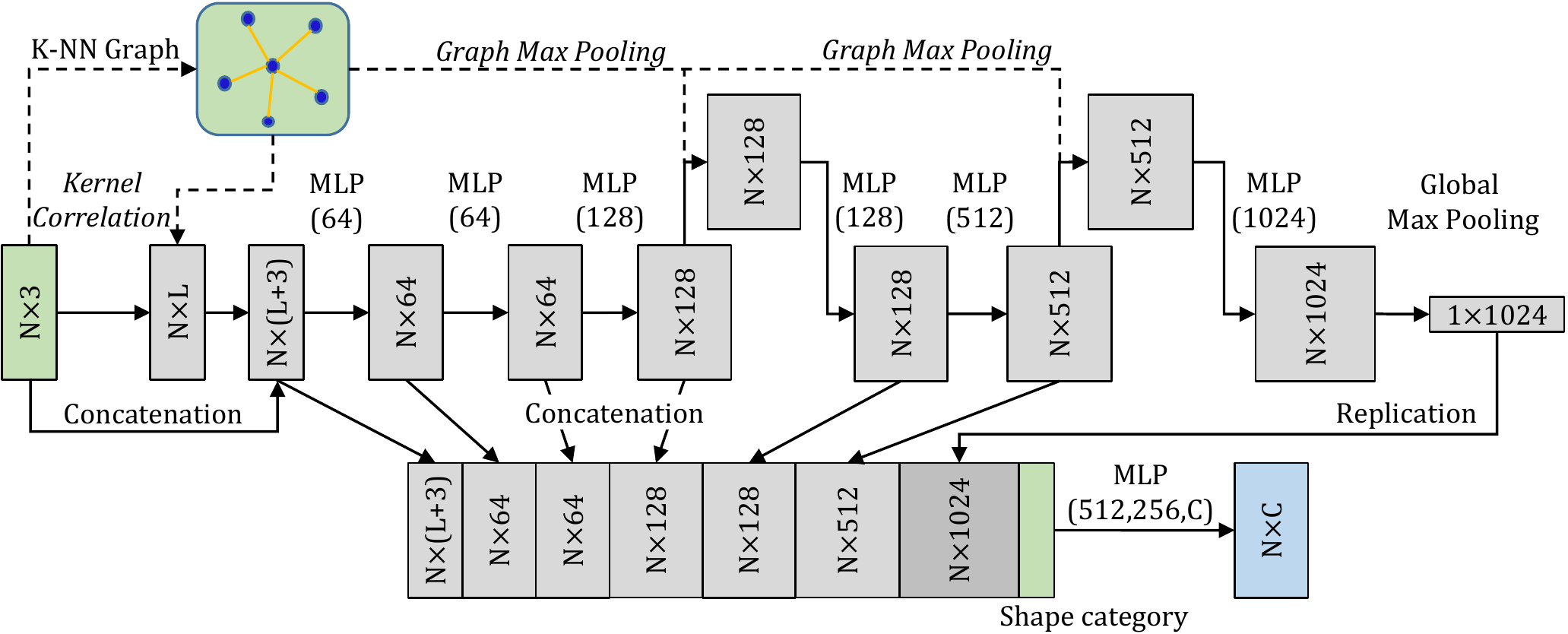}   
   \caption{Segmentation network.}       
   \label{fig:arch-segmentation}    
  \end{subfigure}
\caption{\textbf{Our KCNet architectures.} Local geometric structures are exploited by the front-end kernel correlation layer computing $L$ different affinities between each data point's $K$ nearest neighbor points and $L$ point-set kernels, each kernel containing $M$ learnable 3D points. The resulting responses are concatenated with original 3D coordinates. Local feature structures are later exploited by graph pooling layers, sharing a same graph per 3D object instance constructed offline from each point's 3D Euclidean neighborhood. ReLU is used in each layer without Batchnorm. Dropout layers are used for the last MLPs. Other operations and base network architectures are similar to PointNet~\cite{qi2017pointnet} for both shape classification and part segmentation. Solid arrows indicate forward operation with backward propagation, while dashed arrows mean no backward propagation. Green boxes are input data, gray are intermediate variables, and blue are network predictions. We name our networks as KCNet for short. Best viewed in color.}
\label{fig:architecture}   
\end{figure*}

Nevertheless, influenced by the success of convolutional networks for images, many works have focused on 3D voxels, i.e., regular 3D grids converted from point clouds prior to the learning process. Only then do the 3D convolutional networks learn to extract features from voxels \cite{wu20153d, maturana2015voxnet, qi2016volumetric, li2016fpnn, brock2016generative, dai2017scannet}. 
However, to avoid the intractable computation time complexity and memory consumptions, such methods usually work on a small spatial resolution only, which results in quantization artifacts and difficulties to learn fine details of geometric structures, except for a few recent improvements using Octree~\cite{riegler2017octnet,Wang2017OCNN}.

Different from convolutional nets, PointNet~\cite{qi2017pointnet} provides an effective and simple architecture to directly learn on point sets by firstly computing individual point features from per-point Multi-Layer-Perceptron (MLP) and then aggregating all features as a global presentation of a point cloud.
While achieving state-of-the-art results in different 3D semantic learning tasks, the direct aggregation from per-point features to a global feature suggests that PointNet does not take full advantage of a point's local structure to capture fine-grained patterns:
a per-point MLP output only encodes roughly the existence of a 3D point in a certain nonlinear partition of the 3D space.
A more discriminative representation is expected if the MLP can encode not only ``whether'' a point exists, but also ``what type of'' (e.g., corner vs. planar, convex vs. concave, etc.) a point exists in the non-linear 3D space partition. Such ``type'' information has to be learned from the point's local neighborhood on the 3D object surface, which is the main motivation of this paper.

Attempting to address the above issue, PointNet++~\cite{qi2017pointnetplusplus} propose to segment a point set into smaller clusters, send each through a small PointNet, and repeat such a process iteratively for higher-dimensional feature point sets, which leads to a complicated architecture with reduced speed. Thus, we try to explore from a different direction: is there any efficient learnable local operations with clear geometric interpretations to help directly augment and improve the original PointNet while maintaining its simple architecture?

To address this question, we focus on improving PointNet using two new operations to exploit local geometric and feature structures, as depicted in Figure~\ref{fig:architecture}, regarding two classic supervised representation learning tasks on 3D point clouds. Our contributions are summarized as follows:
\begin{itemize}[noitemsep,nolistsep]
\item We propose a kernel correlation layer to exploit local geometric structures, with a clear geometric interpretation (see Figure~\ref{fig:visualize_KC} and~\ref{fig:visualize_KC_handcraft}).
\item We propose a graph-based pooling layer to exploit local feature structures to enhance network robustness.
\item Our KCNet efficiently improves point cloud semantic learning performances using these two new operations.
\end{itemize}

\section{Related Works} \label{related_work}

\subsection{Local Geometric Properties} \label{local_properties}
We will first discuss some local geometric properties frequently used in 3D data and how they lead us to modify kernel correlation as a tool to enable potentially complex data-driven characterization of local geometric structures.

\textbf{Surface Normal.} \label{normal}
As a basic surface property, surface normals are heavily used in many areas including 3D shape reconstruction, plane extraction, and point set registration~\cite{vosselman20013d, ouyang2005normal, vosselman2004recognising, holz2013fast, chen1991object}.
They usually come directly from CAD models or can be estimated by Principal Component Analysis (PCA) on data covariance matrix of neighboring points as the minimum variance direction~\cite{hoppe1992surface}.
Using per-point surface normal in PointNet corresponds to modeling a point's local neighborhood as a plane, which is shown in~\cite{qi2017pointnet, qi2017pointnetplusplus} to improve performances comparing with only 3D coordinates.
This meets our previous expectation that a point's ``type'' along with its positions should enable better representation.
Yet, this also leads us to a question:
since normals can be estimated from 3D coordinates (not like colors or intensities), then why PointNet with only 3D coordinate input cannot learn to achieve the same performance?
We believe it is due to the following: 1) the per-point MLP cannot capture neighboring information from just 3D coordinates, and 2) global pooling either cannot or is not efficient enough to achieve that.

\textbf{Covariance Matrix.} \label{covariance}
A second-order description of a local neighborhood is through data covariance matrix, which has also been widely used in cases such as plane extraction, curvature estimation~\cite{holz2013fast,feng2014fast} along with normals.
Following the same line of thought from normals, the information provided by the local data covariance matrix is actually richer than normals as it models the local neighborhood as an ellipsoid, which includes lines and planes in rank-deficient cases.
We also observe empirically that it is better than normals for semantic learning.

However, both surface normals and covariance matrices can be seen as handcrafted and limited descriptions of local shapes, because point sets of completely different shapes can share a similar data covariance matrix.
Naturally, to improve performances of both 3D semantic shape classification of fine-grained categories and 3D semantic segmentation, more detailed analysis of each point's local neighborhood is needed.
Although PointNet++~\cite{qi2017pointnetplusplus} is one direct way to learn more discriminative descriptions, it might not be the most efficient solution.
Instead, we would like to find a learnable local description that is efficient, simple, and has a clear geometric interpretation just as the above two handcrafted ones, so it can be directly plugged into the original elegant PointNet architecture.

\textbf{Kernel Correlation.} \label{kernel_correlation}
Another widely used description is the similarity. For images, convolution (often implemented as cross-correlation) can quantify the similarity between an input image and a convolution kernel~\cite{lecun1998gradient}. Yet in face of the aforementioned challenge of defining convolution on point clouds, how can we measure the correlation between two point sets?
This question leads us to kernel correlation~\cite{tsin2004correlation,jian2011robust} as such a tool.
It has been shown that kernel correlation as a function of pairwise point distance is an efficient way to measure geometric affinity between 2D/3D point sets and has been used in point cloud registration and feature correspondence problems~\cite{scott1991algorithm, tsin2004correlation, jian2011robust}.
For registration, in particular, a source point cloud is transformed to best match a reference one by iteratively refining a rigid/non-rigid transformation between the two to maximize their kernel correlation response.
Thus, we propose the kernel correlation layer to treat local neighboring points and a learnable point-set kernel as the source and reference respectively, which is further detailed in Section~\ref{method:KC}.

\subsection{Deep Learning on Point Clouds} \label{neighbor_graph}

Recently, deep learning on 3D input data, especially point clouds, attracts increasing research attention.
There exist four groups of approaches: volumetric-based, patch-based, graph-based and point-based.
\emph{Volumetric-based approach} partitions the 3D space into regular voxels and apply 3D convolution on the voxels \cite{wu20153d, maturana2015voxnet, qi2016volumetric, li2016fpnn, brock2016generative, dai2017scannet}.
However, volumetric representation requires a high memory and computational cost to increase spatial resolution.
Octree-based and kd-tree based networks have been introduced recently, but they could still suffer from the memory efficiency problem \cite{riegler2017octnet, Wang2017OCNN, klokov2017escape}.
\emph{Patch-based approach} parameterizes 3D surface into local patches and apply convolution over these patches \cite{masci2015geodesic, bronstein2017geometric}.
The advantage of this approach is the invariance to surface deformations. Yet it is non-trivial to generalize from mesh to point clouds~\cite{yang2018foldingnet}.
\emph{Graph-based approach} characterizes point clouds by graphs.
Naturally, graph representation is flexible to irregular or even non-Euclidean data such as point clouds, user data on a social network, and gene data \cite{kempe2003maximizing, duvenaud2015convolutional, kipf2016semi, niepert2016learning,  niepert2016learning, atwood2016diffusion, monti2016geometric}.
Therefore, a graph e.g. a connectivity graph or a polygon mesh can be used to represent a 3D point cloud, convert to the spectral representation and apply convolution in spectral domain \cite{bruna2013spectral, henaff2015deep, edwards2016graph, defferrard2016convolutional, kipf2016semi, levie2017cayleynets}.
Another study also investigates convolution over edge attributes in the neighborhood of a vertex from graphs built on point clouds \cite{simonovsky2017dynamic}.
\emph{Point-based approach} such as PointNet directly operates on point clouds, with spatial features learned for each point, and global features obtained by aggregating over point features through max-pooling \cite{qi2017pointnet}.
PointNet is simple yet efficient for the applications of shape classification and segmentation.
However, global aggregation without explicitly considering local structures misses the opportunity to capture fine-grained patterns and suffers from sensitivity to noises.
To further extend PointNet to local structures, we use a simple graph-based network: we construct k nearest neighbor graphs (KNNG) to utilize the neighborhood information for kernel correlation and to recursively conduct the max-pooling operations in each node’s neighborhood, with the insight that local points share similar geometric structures.
KNNG is usually used to establish local connectivity information, in the applications of point cloud on surface detection, 3D object recognition, 3D object segmentation and compression~\cite{golovinskiy2009shape, strom2010graph, thanou2016graph}.

\section{Method} \label{method}
We now explain the details of learning local structures over point neighborhoods by 1) kernel correlation that measures the geometric affinity of point sets, and 2) a KNNG that propagates local features between neighboring points.
Figure \ref{fig:architecture} illustrates our full network architectures.

\subsection{Learning on Local Geometric Structure} \label{method:KC}

As mentioned earlier, in our network's front-end, we take inspiration from kernel correlation based point cloud registration and treat a point's local neighborhood as the source, and a set of learnable points, i.e., a kernel, as the reference that characterizes certain types of local geometric structures/shapes.
We modify the original kernel correlation computation by allowing the reference to freely adjust its shape (kernel point positions) through backward propagation.
Note the change of perspective here compared with point set registration: we want to learn template/reference shapes through a free per-point transformation, instead of using a fixed template to find an optimal transformation between source and reference point sets.
In this way, a set of learnable kernel points is analogous to a convolutional kernel, which activates to points only in its joint neighboring regions and captures local geometric structures within this receptive field characterized by the kernel function and its kernel width.
Under this setting, the learning process can be viewed as finding a set of reference/template points encoding the most effective and useful local geometric structures that lead to the best learning performance jointly with other parameters in the network.

Specifically, we adapt ideas of the \textit{Leave-one-out Kernel Correlation} (LOO-KC) and the \textit{multiply-linked} registration cost function in~\cite{tsin2004correlation} to capture local geometric structures of a point cloud. Let us define our kernel correlation (KC) between a point-set kernel $\bm{\kappa}$ with $M$ learnable points and the current anchor point $\x_i$ in a point cloud of $N$ points as:
\begin{equation} \label{equation:KC}
\KC(\bm{\kappa}, \x_i) = \frac{1}{|\mathcal{N}(i)|} \sum_{m=1}^{M} \sum_{n\in\mathcal{N}(i)} {\K}_{\sigma}(\bm{\kappa}_m, \x_n-\x_i),
\end{equation}
where $\bm{\kappa}_m$ is the m-th learnable point in the kernel, $\mathcal{N}(i)$ is the neighborhood index set of the anchor point $\x_i$, and $\x_n$ is one of $\x_i$'s neighbor point.
${\K}_{\sigma}(\cdot, \cdot): \mathbb{\Re}^{D}$ $\times$ $\mathbb{\Re}^{D}$ $\rightarrow$ $\mathbb{\Re}$
is any valid kernel function ($D=2\text{ or }3$ for 2D or 3D point clouds).
To efficiently store the local neighborhood of points, we pre-compute a KNNG by considering each point as a vertex, with edges connecting only nearby vertices.

Following~\cite{tsin2004correlation}, without loss of generality, we choose the Gaussian kernel in this paper:
\begin{equation} \label{equation:gaussian-kernel}
{\K}_\sigma(\mathbf{k}, \bm{\delta}) = \exp\left( - \frac{||\mathbf{k} - \bm{\delta}||^{2}}{2 \sigma^2} \right)
\end{equation}
where $||\cdot||$ is the Euclidean distance between two points and $\sigma$ is the kernel width that controls the influence of distance between points.
One nice property of Gaussian kernel is that it decays exponentially as a function of the distance between the two points, providing a soft-assignment from each kernel point to neighboring points of the anchor point, relaxing from the non-differentiable hard-assignment in ordinary ICP.
Our KC encodes pairwise distance between kernel points and neighboring data points and increases as two point sets become similar in shape, hence it can be clearly interpreted as a geometric similarity measure, and is invariant under translation.
Note the importance of choosing kernel width here, since either a too large or a too small $\sigma$ will lead to undesired performances (see Table~\ref{table:hyper-parameters}), similar to the same issue in kernel density estimation.
Fortunately, for 2D or 3D space in our case, this parameter can be empirically chosen as the average neighbor distance in the neighborhood graphs over all training point clouds.

\begin{figure}[t]
	\begin{tabular}{@{}c@{}c@{}c@{}c@{}c@{}c@{}}
		\includegraphics[width=0.16\columnwidth,height=1.3cm,keepaspectratio]{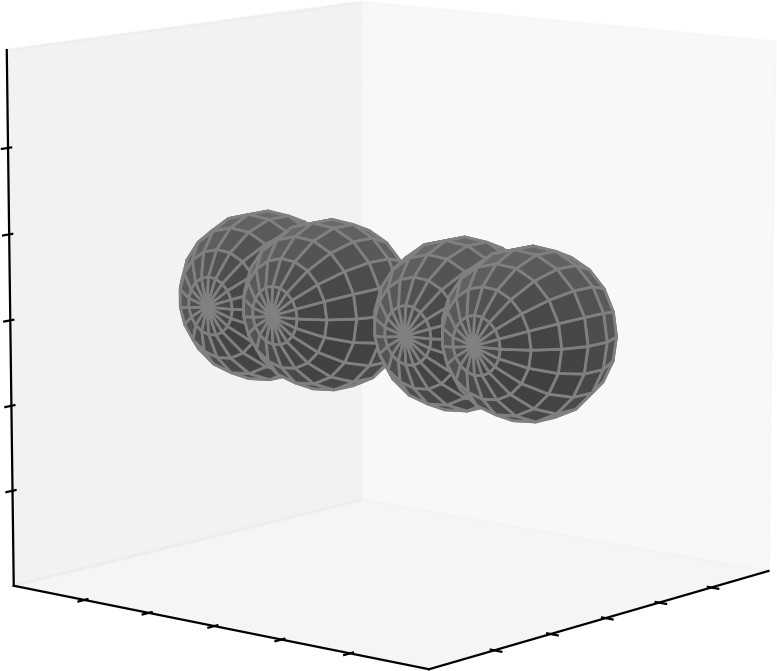} &
		\includegraphics[width=0.16\columnwidth,height=1.3cm,keepaspectratio]{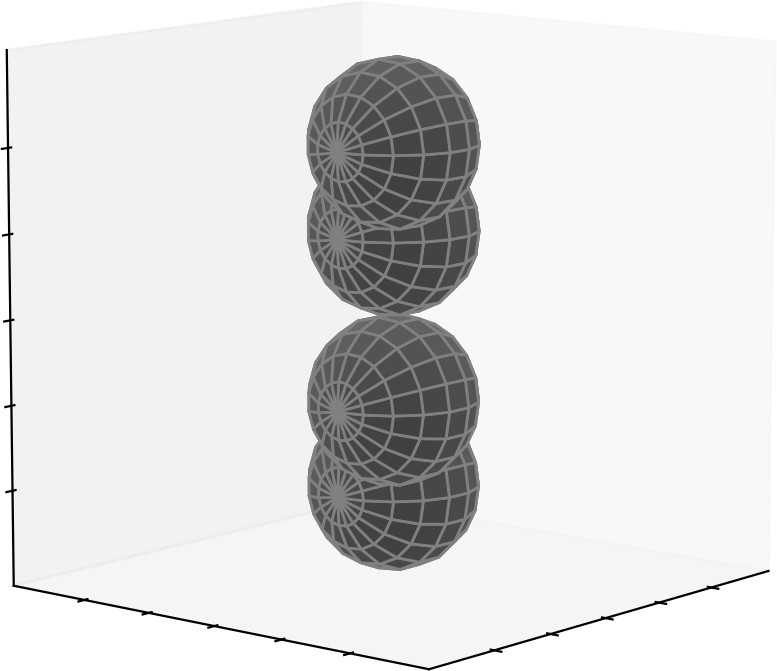} &
		\includegraphics[width=0.16\columnwidth,height=1.3cm,keepaspectratio]{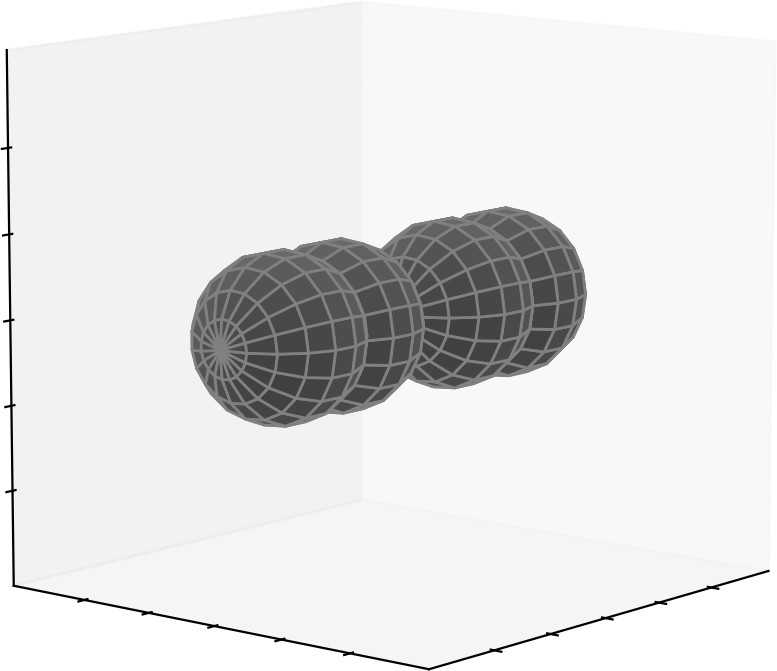} &
		\includegraphics[width=0.16\columnwidth,height=1.3cm,keepaspectratio]{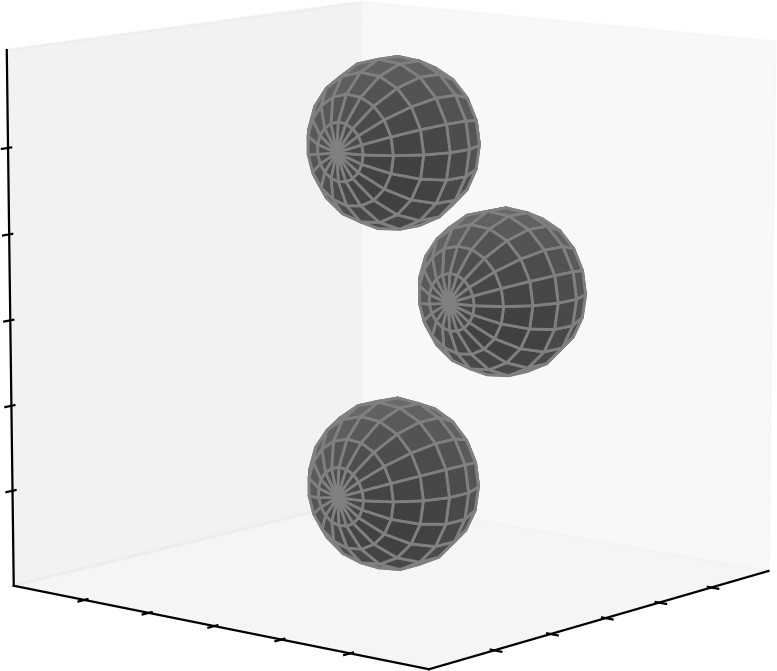} &
		\includegraphics[width=0.16\columnwidth,height=1.3cm,keepaspectratio]{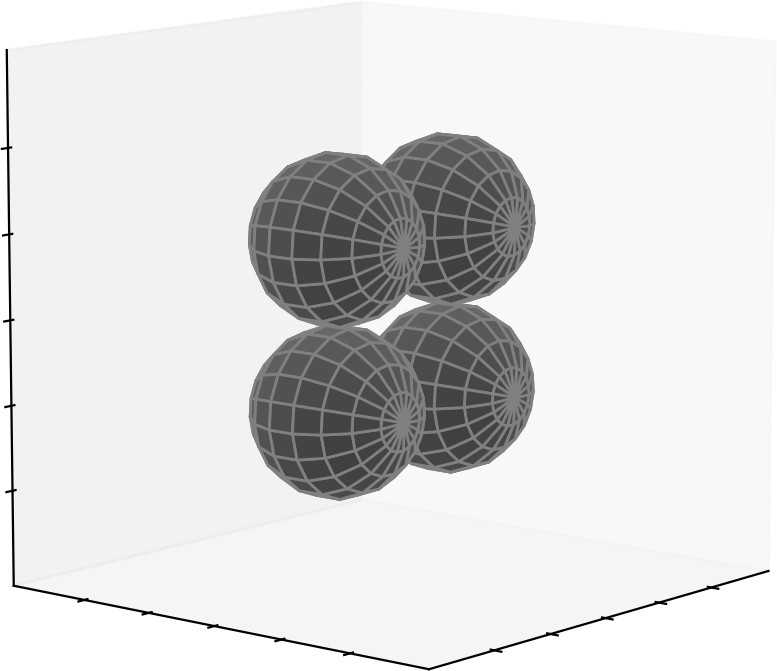} &
		\includegraphics[width=0.16\columnwidth,height=1.3cm,keepaspectratio]{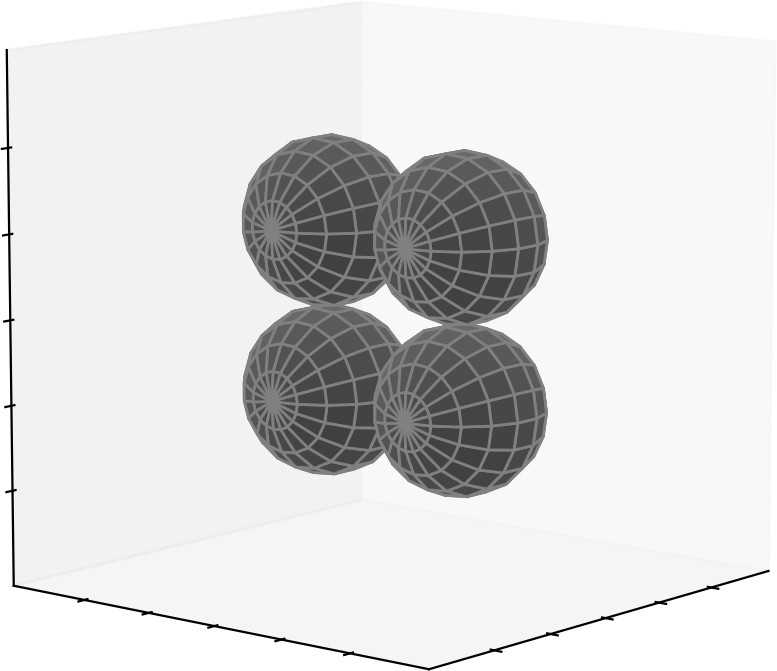} \\    	
		
		\includegraphics[width=0.16\columnwidth,height=1.3cm,keepaspectratio]{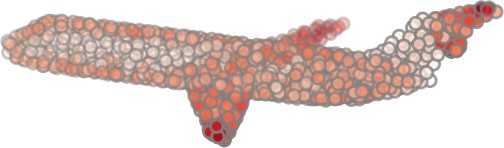} &
		\includegraphics[width=0.16\columnwidth,height=1.3cm,keepaspectratio]{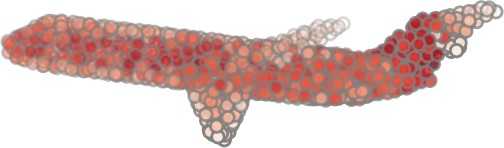} &
		\includegraphics[width=0.16\columnwidth,height=1.3cm,keepaspectratio]{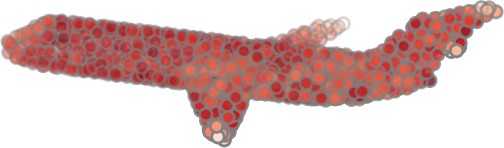} &
		\includegraphics[width=0.16\columnwidth,height=1.3cm,keepaspectratio]{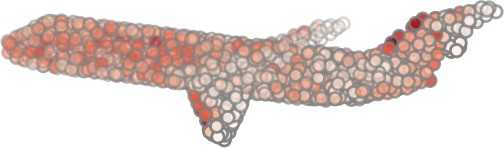} &
		\includegraphics[width=0.16\columnwidth,height=1.3cm,keepaspectratio]{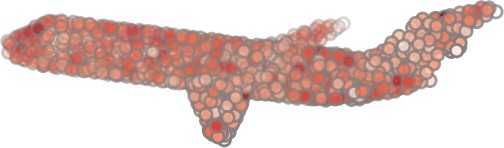} &
		\includegraphics[width=0.16\columnwidth,height=1.3cm,keepaspectratio]{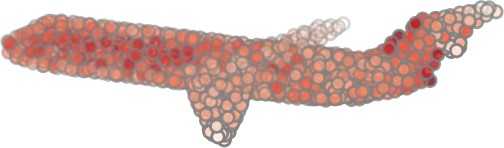} \\    	
		
		\includegraphics[width=0.16\columnwidth,height=1.3cm,keepaspectratio]{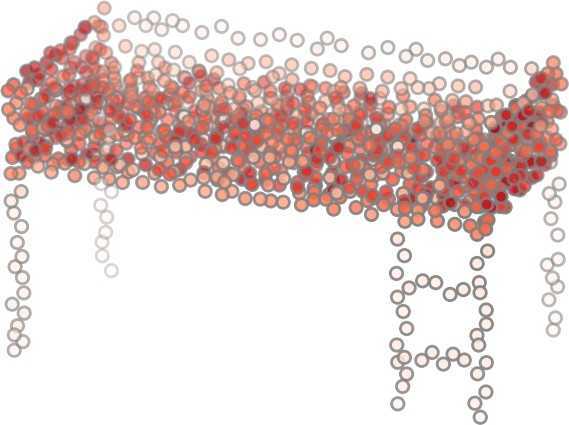} &
		\includegraphics[width=0.16\columnwidth,height=1.3cm,keepaspectratio]{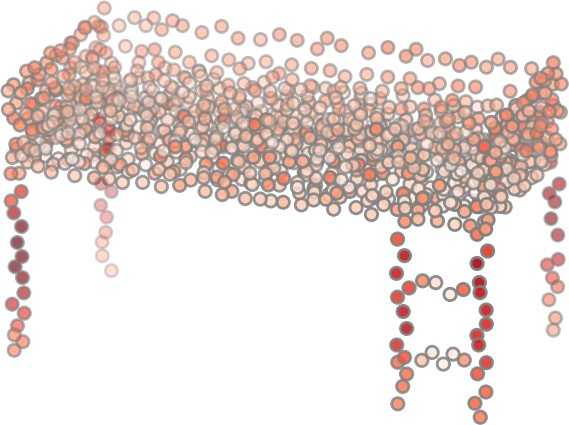} &
		\includegraphics[width=0.16\columnwidth,height=1.3cm,keepaspectratio]{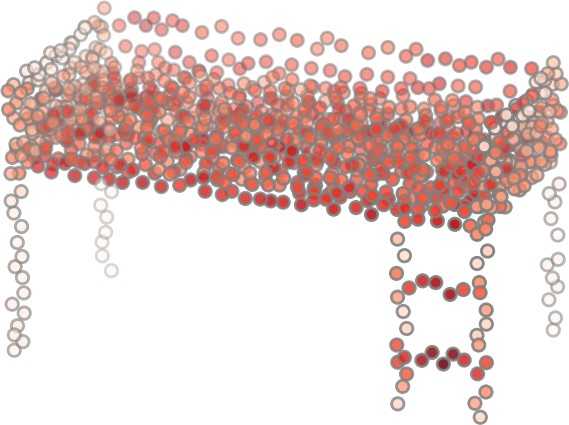} &
		\includegraphics[width=0.16\columnwidth,height=1.3cm,keepaspectratio]{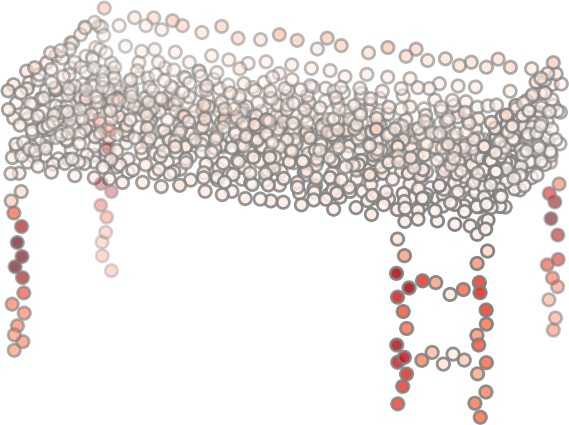} &
		\includegraphics[width=0.16\columnwidth,height=1.3cm,keepaspectratio]{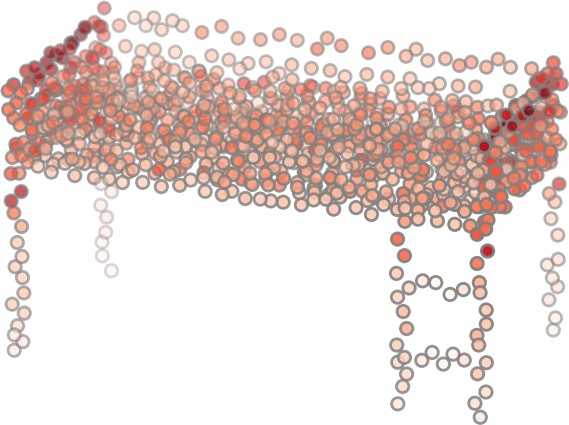} &
		\includegraphics[width=0.16\columnwidth,height=1.3cm,keepaspectratio]{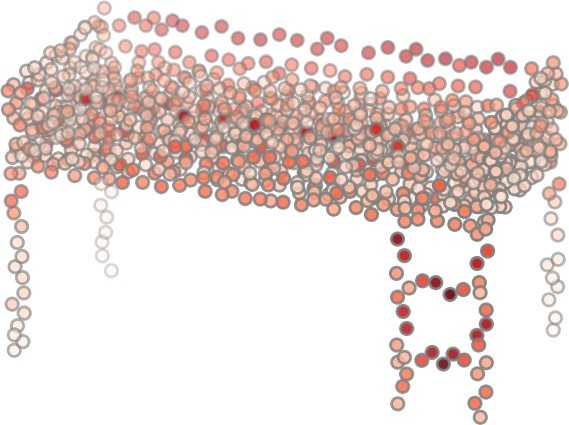} \\    	
		
		\includegraphics[width=0.16\columnwidth,height=1.3cm,keepaspectratio]{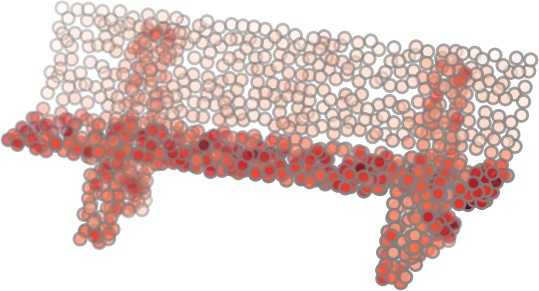} &
		\includegraphics[width=0.16\columnwidth,height=1.3cm,keepaspectratio]{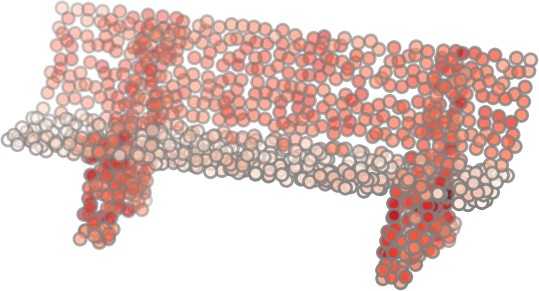} &
		\includegraphics[width=0.16\columnwidth,height=1.3cm,keepaspectratio]{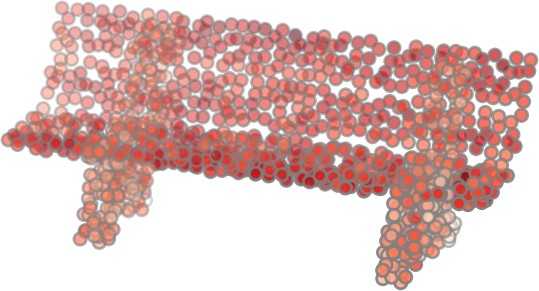} &
		\includegraphics[width=0.16\columnwidth,height=1.3cm,keepaspectratio]{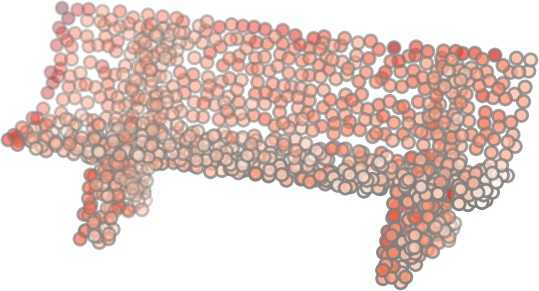} &
		\includegraphics[width=0.16\columnwidth,height=1.3cm,keepaspectratio]{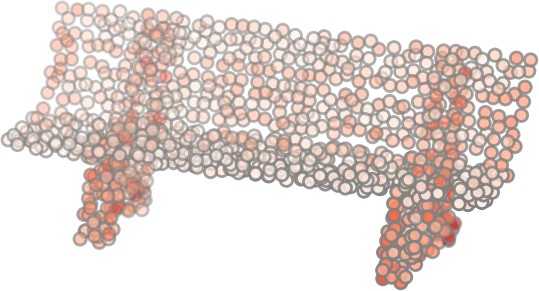} &
		\includegraphics[width=0.16\columnwidth,height=1.3cm,keepaspectratio]{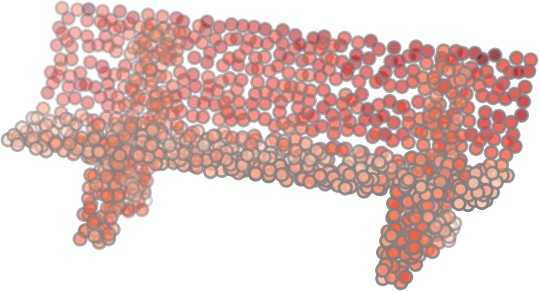} \\    
		
		\includegraphics[width=0.16\columnwidth,height=1.3cm,keepaspectratio]{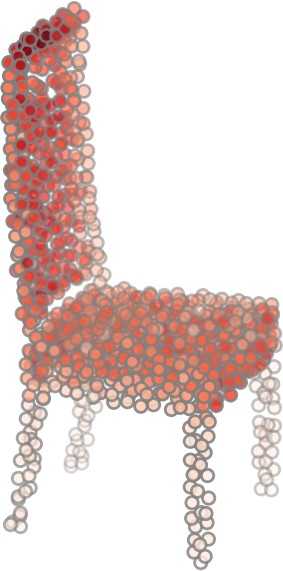} &
		\includegraphics[width=0.16\columnwidth,height=1.3cm,keepaspectratio]{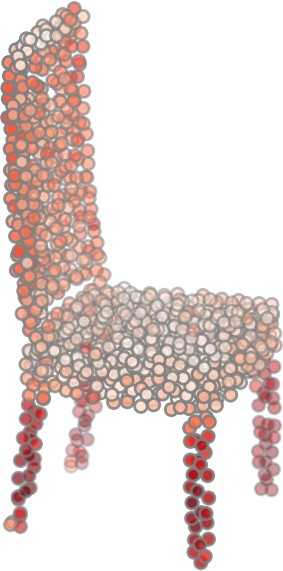} &
		\includegraphics[width=0.16\columnwidth,height=1.3cm,keepaspectratio]{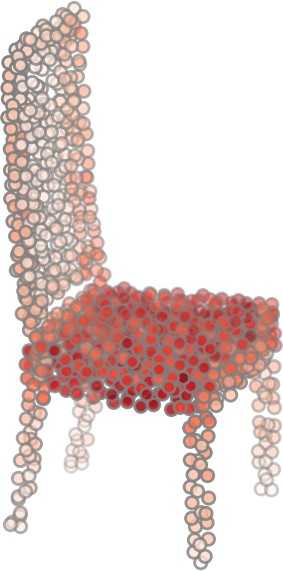} &
		\includegraphics[width=0.16\columnwidth,height=1.3cm,keepaspectratio]{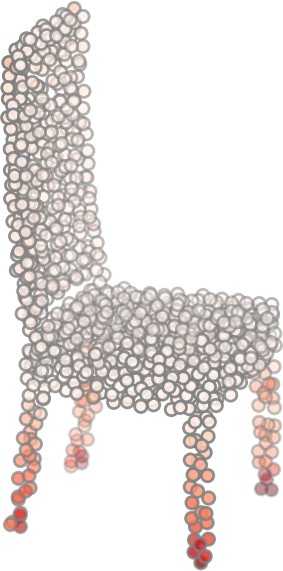} &
		\includegraphics[width=0.16\columnwidth,height=1.3cm,keepaspectratio]{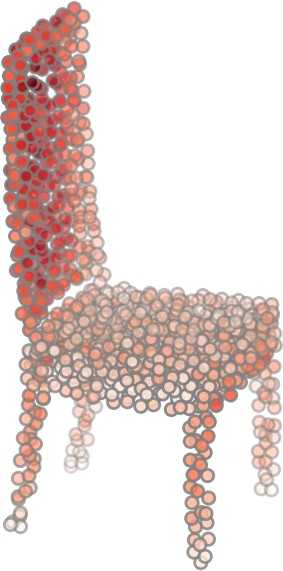} &
		\includegraphics[width=0.16\columnwidth,height=1.3cm,keepaspectratio]{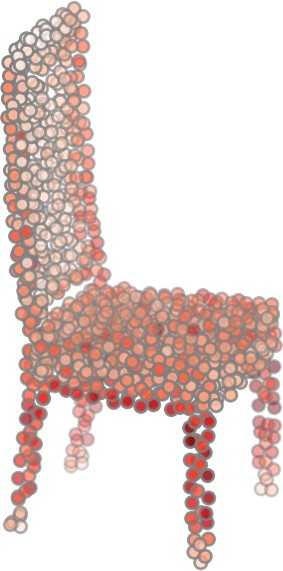} \\   
		
		\includegraphics[width=0.16\columnwidth,height=1.3cm,keepaspectratio]{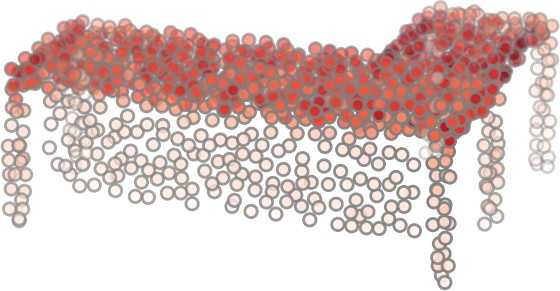} &
		\includegraphics[width=0.16\columnwidth,height=1.3cm,keepaspectratio]{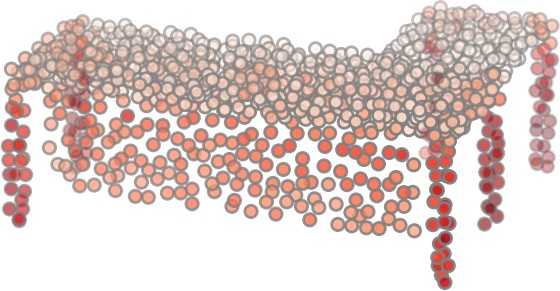} &
		\includegraphics[width=0.16\columnwidth,height=1.3cm,keepaspectratio]{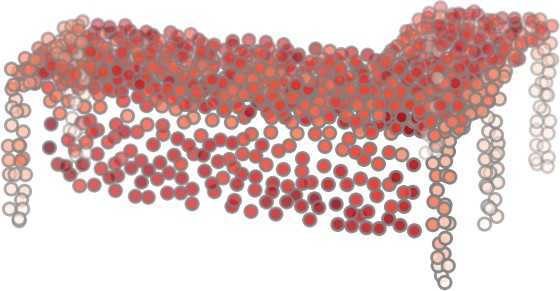} &
		\includegraphics[width=0.16\columnwidth,height=1.3cm,keepaspectratio]{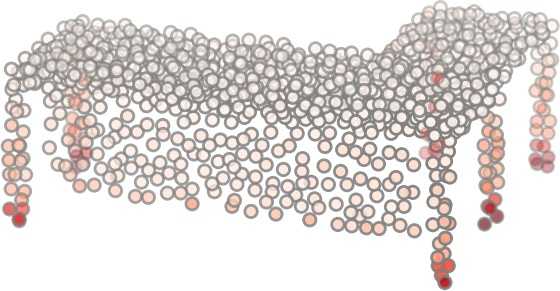} &
		\includegraphics[width=0.16\columnwidth,height=1.3cm,keepaspectratio]{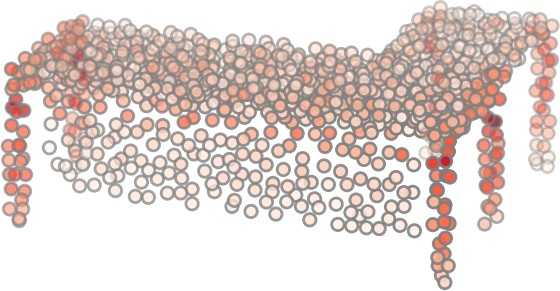} &
		\includegraphics[width=0.16\columnwidth,height=1.3cm,keepaspectratio]{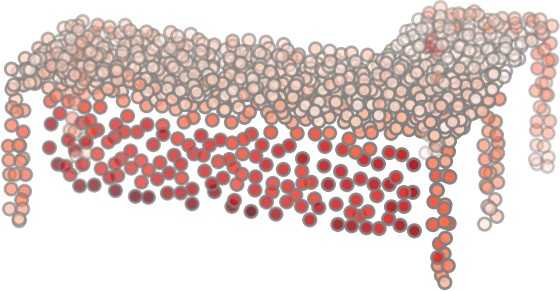} \\   	
	\end{tabular}
	\caption{Visualization of handcrafted (linear, planar, and curved) kernels and responses, similar to Figure~\ref{fig:visualize_KC}. \label{fig:visualize_KC_handcraft}} \vspace{-4mm}
\end{figure}

To complete the description of the proposed new learnable layer, given 1) $\mathcal{L}$ as the network loss function, 2) its derivative w.r.t. each point $\x_i$'s KC response $d_i=\frac{\partial\mathcal{L}}{\partial \KC(\bm{\kappa}, \x_i)}$ propagated back from top layers, we provide the back-propagation equation for each kernel point $\bm{\kappa}_m$ as:
\begin{equation} \label{equation:KC-backprop}
\frac{\partial\mathcal{L}}{\partial\bm{\kappa}_m} = \sum_{i=1}^{N} \alpha_i d_i \Big[ \sum_{n\in\mathcal{N}(i)} \mathbf{v}_{m,i,n} \exp(-\frac{||\mathbf{v}_{m,i,n}||^{2}}{2\sigma^{2}}) \Big],
\end{equation}
where point $x_i$'s normalizing constant $\alpha_i=\frac{-1}{|\mathcal{N}(i)|\sigma^2}$, and the local difference vector $\mathbf{v}_{m,i,n}=\bm{\kappa}_m+\x_i-\x_n$.

Although originates from LOO-KC in~\cite{tsin2004correlation}, our KC operation is different: 1) unlike LOO-KC as a compactness measure between a point set and one of its element point, our KC computes the similarity between a data point's neighborhood and a kernel of learnable points;
and 2) unlike the multiply-linked cost function involving a parameter of a transformation for a fixed template, our KC allows all points in the kernel to freely move and adjust (i.e., no weight decay for $\bm{\kappa}$), thus replacing the template and the transformation parameters as a point-set kernel.

To better understand how KC captures various local geometric structures, we visualize several handcrafted kernels and corresponding KC responses on different objects in Figure~\ref{fig:visualize_KC_handcraft}. Similarly, we visualize several learned kernels from our segmentation network in Figure~\ref{fig:visualize_KC}. Note that we can learn $L$ different kernels in KCNet, where $L$ is a hyper-parameter similar to the number of output channels in convolutional nets.

\subsection{Learning on Local Feature Structure} \label{method:graph}

Our KCNet performs KC only in the network front-end to extract local geometric structure, as shown in Figure~\ref{fig:architecture}.
For computing KC, to efficiently store the local neighborhood of points, we build a KNNG by considering each point as a vertex, with edges connecting only nearby points.
This graph is also useful for exploiting local feature structures in deeper layers.
Inspired by the ability of convolutional nets to locally aggregate features and gradually increase receptive fields via multiple pooling layers, we use recursive feature propagation and aggregation along edges of the very same 3D neighborhood graph for KC, to exploit local feature structures in the top layers.

Our key insight is that neighbor points tend to have similar geometric structures and hence propagating features through neighborhood graph helps to learn more robust local patterns.
Note that we specifically avoid changing this neighborhood graph structure in top layers, which is also analogous to convolution on images:
each pixel's spatial ordering and neighborhoods remain unchanged even when feature channels of input image expand greatly in top convolutional layers.

Specifically, let $\mathbf{X} \in \mathbb{\Re}^{N \times K}$ represent input to the graph pooling layer, and the KNNG has an adjacency matrix $\mathbf{W} \in \mathbb{\Re}^{N \times N}$ where $\mathbf{W}(i,j)=1$ if there exists an edge between vertex $i$ and $j$, and $\mathbf{W}(i,j)=0$ otherwise.
It is intuitive that neighboring points forming local surface often share similar feature patterns.
Therefore, we aggregate features of each point within its neighborhood by a graph pooling operation:
\begin{equation} \label{equation:graph-operation}
\mathbf{Y} = \mathbf{P}\mathbf{X},
\end{equation}
which can be implemented as average or max pooling.

Graph average pooling averages a point's features over its neighborhood by using $\mathbf{P} \in \mathbb{\Re}^{N\times N}$ in~\eqref{equation:graph-operation} as a normalized adjacency matrix:
\begin{equation}
\label{equation:normalized-adjancency}
\mathbf{P} = \mathbf{D}^{-1} \mathbf{W},
\end{equation}
where $\mathbf{D} \in \mathbb{\Re}^{N\times N}$ is the degree matrix with $(i,j)$-th entry $d_{i,j}$ defined as:
\begin{equation}
\label{equation:degree-matrix}
d_{i,j} =
\begin{cases}
deg(i), & if \quad i=j \\
0, & otherwise
\end{cases}
\end{equation}
where $deg(i)$ is the degree of vertex $i$ counting the number of vertices connected to vertex $i$.

Graph max pooling (GM) takes maximum features over the neighborhood of each vertex, independently operated over each of the $K$ dimensions.
This can be simply computed by replacing the ``$+$'' operator in the matrix multiplication in~\eqref{equation:graph-operation} with a ``$\max$'' operator.
Thus the $(i, k)$-th entry of output $\mathbf{Y}$ is:
\begin{equation}
\mathbf{Y}(i, k) = \max_{n\in\mathcal{N}(i)} \mathbf{X}(n, k),
\end{equation}
where $\mathcal{N}(i)$ indicates the neighbor index set of point $\mathbf{X}_i$ computed from $\mathbf{W}$.

A point's local signature is then obtained by graph max or average pooling.
This signature can represent the aggregated feature information of the local surface. 
Note the connection of this operation with PointNet++: each point $i$'s local neighborhood is similar to the clusters/segments in PointNet++. This graph operation enables local feature aggregation on the original PointNet architecture.

\section{Experiments}
\label{experiments}
Now we discuss the proposed architectures for 3D shape classification (Section~\ref{experiment:classification}),  part segmentation (Section~\ref{experiment:part-segmentation}), and perform ablation study (Section~\ref{experiment:analysis}).

\subsection{Shape Classification} \label{experiment:classification}
\noindent\\
\textbf{Datasets.} We evaluated our network on both 2D and 3D point clouds. For 2D shape classification, we converted MNIST dataset~\cite{lecun1998gradient} to 2D point clouds. MNIST contains images of handwritten digits with 60,000 training and 10,000 testing images. We transformed non-zero pixels in each image to 2D points, keeping coordinates as input features and normalize them within [-0.5, 0.5]. For 3D shape classification, we evaluated our KCNet on 10-categories and 40-categories benchmarks ModelNet10 and ModelNet40~\cite{wu20153d}, consisting of 4899 and 12311 CAD models respectively. ModelNet10 is split into 3991 for training and 908 for testing. ModelNet40 is split into 9843 for training and 2468 for testing. As in PointNet, to obtain 3D point clouds, we uniformly sampled points from meshes into 1024 points of each object by Poisson disk sampling using MeshLab~\cite{cignoni2008meshlab} and normalized them into a unit ball.

\noindent\\
\textbf{Network Configuration.} As detailed in Figure~\ref{fig:arch-classification}, our KCNet has 9 parametric layers in total. 
The first layer, kernel correlation, takes point coordinates as inputs and outputs local geometric features and concatenated with the point coordinates. 
Then features are passed into the first 2-layer MLP for per-point feature learning.
The graph pooling layer then aggregates the output per-point features into more robust local structure features, which are concatenated with the outputs from the second 2-layer MLP.
Other configurations are similar to the original PointNet, except that 1) ReLU is used after each fully connected layer without Batchnorm (we found it not useful in KCNet and PointNet), and 2) Dropout layers are used for the final fully connected layers with drop ratio 0.5.
We used 16-NN graph for kernel computation and graph max pooling.
$L=32$ sets of kernels were used, in which each kernel had $M=16$ points uniformly initialized within [-0.2, 0.2] and kernel width $\sigma=0.005$.
We trained the network for 400 epochs on a NVIDIA GTX 1080 GPU using our modified Caffe~\cite{jia2014caffe} with ADAM optimizer, initial learning rate 0.001, batch size 64, momentum 0.9, momentum2 0.999, and weight decay  $1\mathrm{e}{-5}$.
No data augmentation was performed.

\noindent\\
\textbf{Results.} Table~\ref{table:mnist} and Table~\ref{table:modelNet} compares our results with several recent works. In MNIST digit classification, KCNet reaches comparable results obtained with ConvNets. In ModelNet40 shape classification, our method achieves competitive performance with 3.8\% and 1.8\% higher accuracy than PointNet-vanilla (meaning without T-nets) and PointNet respectively~\cite{qi2017pointnet}, and is slightly better (0.3\%) than PointNet++~\cite{qi2017pointnetplusplus}. Table~\ref{table:time-model-size} summarizes required number of parameters and forward time of different networks.
Note KCNet achieves better or comparable accuracy and computes more efficiently than~\cite{qi2017pointnetplusplus,klokov2017escape} with fewer parameters.

\begin{table}[!t]
    \centering
    \begin{tabular}{l|c}
    \hline
    Method & Accuracy (\%) \\ \hline
    LeNet5 \cite{lecun1998gradient} & 99.2 \\
    PointNet (vanilla) \cite{qi2017pointnet} & 98.7 \\
    PointNet \cite{qi2017pointnet} & 99.2 \\
    PointNet++ \cite{qi2017pointnetplusplus} & \textbf{99.5} \\ \hline
    KCNet (ours) & 99.3 \\ \hline
    \end{tabular}\vspace{-2mm}
    \caption{MNIST digit classification. \label{table:mnist}} \vspace{-1mm}
\end{table}

\begin{table}[!t]
    \centering
    \begin{tabular}{l|c c}
    \hline
    Method & MN10 & MN40 \\ \hline
    MVCNN \cite{su2015multi} & - & 90.1 \\
    VRN Ensemble \cite{brock2016generative} & \textbf{97.1} & \textbf{95.5} \\ \hline
    ECC \cite{simonovsky2017dynamic} & 90.0 & 83.2 \\
    PointNet (vanilla) \cite{qi2017pointnet} & - & 87.2 \\
    PointNet \cite{qi2017pointnet} & -  & 89.2 \\
    PointNet++ \cite{qi2017pointnetplusplus} & - & 90.7 \\
    Kd-Net(depth 10) \cite{klokov2017escape} & 93.3 & 90.6 \\
    Kd-Net(depth 15) \cite{klokov2017escape}  & 94.0 & 91.8 \\ \hline
    KCNet (ours) & 94.4 & 91.0 \\ \hline
    \end{tabular} \vspace{-2mm}
    \caption{ModelNet shape classification comparisons of accuracy of proposed network with state-of-the-art. Our KCNet has competitive performance on both ModelNet10 and ModelNet40. Note that MVCNN \cite{su2015multi} and VRN Ensemble \cite{brock2016generative} take image and volume as inputs, while rest of the models take point clouds as inputs. \label{table:modelNet}} \vspace{-1mm}
\end{table}

\begin{table}[!t]
    \centering
    \begin{tabular}{l@{}|cc}
    \hline
     Method & \#params (M) & Fwd. time (ms) \\ \hline
     PointNet(vanilla)\cite{qi2017pointnetplusplus} & \textbf{0.8} & \textbf{11.6} \\
     PointNet\cite{qi2017pointnetplusplus}  & 3.5 & 25.3  \\
     PointNet++(MSG)\cite{qi2017pointnetplusplus} & 1.0 & 163.2 \\
     Kd-Net (depth 10) & 2.0 & - \\ \hline
     KCNet ($M=16$) & 0.9 & 18.5 \\
     KCNet ($M=3$)  & 0.9 & 12.0 \\ \hline
    \end{tabular} \vspace{-1mm}
    \caption{Model size and inference time. "M" stands for million. Networks were tested on a PC with a single NVIDIA GTX 1080 GPU and an Intel i7-8700@3.2 GHz 12 cores CPU. Other settings are the same as in~\cite{qi2017pointnetplusplus}. \label{table:time-model-size}} \vspace{-3mm}
    
\end{table}

\subsection{Part Segmentation}
\label{experiment:part-segmentation}
Part segmentation is an important task that requires accurate segmentation of complex shapes with delicate structures. We used the network illustrated in Figure~\ref{fig:arch-segmentation} to predict the part label of each point in a 3D point cloud object.

\noindent\\
\textbf{Datasets.} We evaluated KCNet for part segmentation on ShapeNet part dataset~\cite{yi2016scalable}. There are 16,881 shapes of 3D point cloud objects from 16 shape categories, with each point in an object corresponds to a part label (50 parts in total, and non-overlapping across shape categories). On average each object consists of less than 6 parts and the highly imbalanced data makes the task quite challenging. We used the same strategy as in Section~\ref{experiment:classification} to uniformly sample 2048 points for each CAD object.
We used the official train test split following \cite{qi2017pointnetplusplus}.

\begin{table*}
    \setlength{\tabcolsep}{0.15em}
    \begin{tabular}{@{} c|c|c|ccccccccccccccccc} \hline
    \multicolumn{1}{l}{} & \multicolumn{1}{|c|}{Cat.} & \multicolumn{1}{c|}{Ins.} & \multicolumn{1}{c}{aero} & \multicolumn{1}{c}{bag} & \multicolumn{1}{c}{cap} & \multicolumn{1}{c}{car} & \multicolumn{1}{c}{chair} & \multicolumn{1}{c}{ear} &  \multicolumn{1}{c}{guitar} & \multicolumn{1}{c}{knife} & \multicolumn{1}{c}{lamp} & \multicolumn{1}{c}{laptop} & \multicolumn{1}{c}{motor} & \multicolumn{1}{c}{mug} & \multicolumn{1}{c}{pistol} & \multicolumn{1}{c}{rocket} & \multicolumn{1}{c}{skate} & \multicolumn{1}{c}{table} \\
    \multicolumn{1}{c}{} &\multicolumn{1}{|c|}{mIoU} & \multicolumn{1}{c|}{mIoU} &  \multicolumn{1}{c}{} & \multicolumn{1}{c}{} & \multicolumn{1}{c}{} & \multicolumn{1}{c}{} & \multicolumn{1}{c}{} & \multicolumn{1}{c}{phone} & \multicolumn{1}{c}{} & \multicolumn{1}{c}{} & \multicolumn{1}{c}{} & \multicolumn{1}{c}{} & \multicolumn{1}{c}{} & \multicolumn{1}{c}{} & \multicolumn{1}{c}{} & \multicolumn{1}{c}{} &\multicolumn{1}{c}{board} &\multicolumn{1}{c}{}  \\
    \midrule
    \# shapes & & & 2690 & 76 & 55 & 898 & 3758 & 69 & 787 & 392 & 1547 & 451 & 202 & 184 & 283 & 66 & 152 & 5271 \\ \hline
    PointNet & 80.4 & 83.7 & \textbf{83.4} & 78.7 & 82.5 & 74.9 & 89.6 & 73.0 & \textbf{91.5}  & 85.9 & 80.8 & 95.3 & 65.2 & 93.0 & 81.2 & 57.9 & 72.8 & 80.6 \\
    PointNet++ & 81.9 & \textbf{85.1} & 82.4 & 79.0 & \textbf{87.7} & 77.3 & \textbf{90.8} & 71.8 & 91.0 & 85.9 & 83.7 & 95.3 & \textbf{71.6} & 94.1 & 81.3 & 58.7 & \textbf{76.4} & \textbf{82.6}\\
    Kd-Net & 77.4 & 82.3 & 80.1 & 74.6 & 74.3 & 70.3 & 88.6 & 73.5 & 90.2 & \textbf{87.2} & 81.0 & 94.9 & 57.4 & 86.7 & 78.1 & 51.8 & 69.9 & 80.3 \\ \hline
    KCNet (ours) & \textbf{82.2} & 84.7 & 82.8 & \textbf{81.5} & 86.4 & \textbf{77.6} & 90.3 & \textbf{76.8} & 91.0 & \textbf{87.2} & \textbf{84.5} & \textbf{95.5} & 69.2 & \textbf{94.4} & \textbf{81.6} & \textbf{60.1} & 75.2 & 81.3 \\ \hline
    \end{tabular} \vspace{-2mm}
    \caption{ShapeNet part segmentation results. Average mIoU over instances (Ins.) and categories (Cat.) are reported. \label{table:part-segmentation}}
\end{table*}

\begin{figure*}[ht]
	\begin{tabular}{ccc|ccc|ccc}
		\toprule
		GT & PointNet & Ours & GT & PointNet & Ours & GT & PointNet & Ours \\
		\midrule
		\includegraphics[width=0.09\textwidth,height=1.1cm]{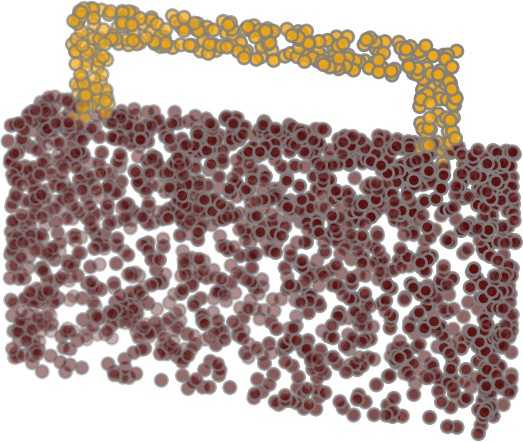} & \includegraphics[width=0.09\textwidth,height=1.1cm]{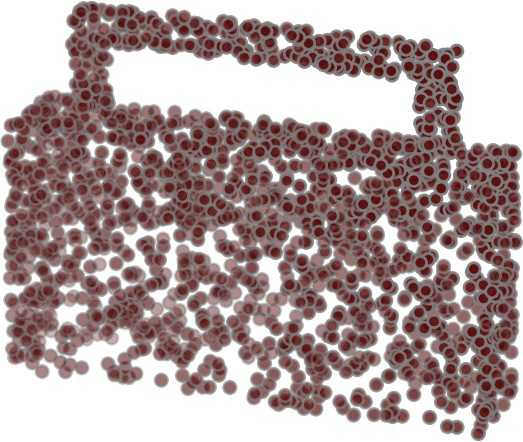} & \includegraphics[width=0.09\textwidth,height=1.1cm]{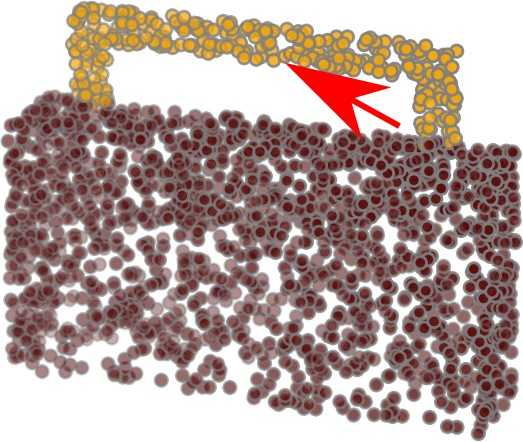} & \includegraphics[width=0.09\textwidth,height=1.1cm]{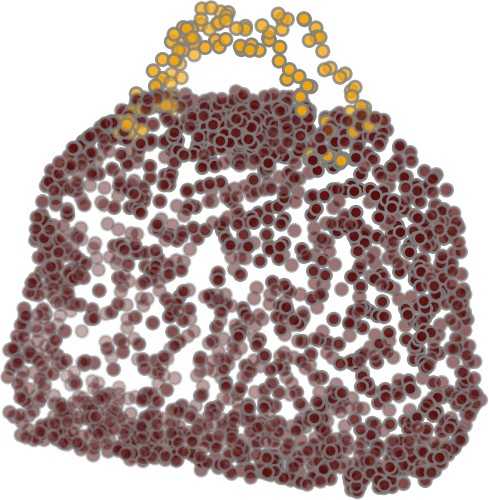} & \includegraphics[width=0.09\textwidth,height=1.1cm]{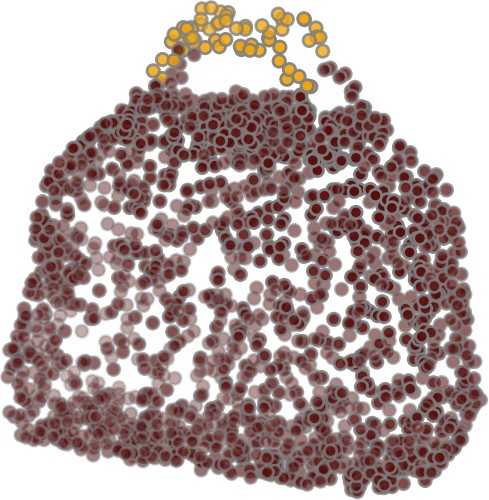}& \includegraphics[width=0.09\textwidth,height=1.1cm]{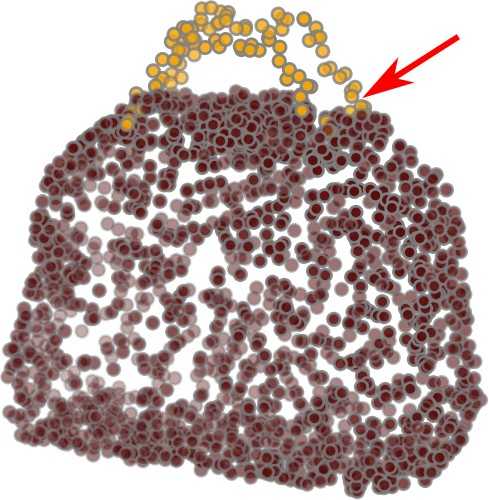} & \includegraphics[width=0.09\textwidth,height=1.1cm]{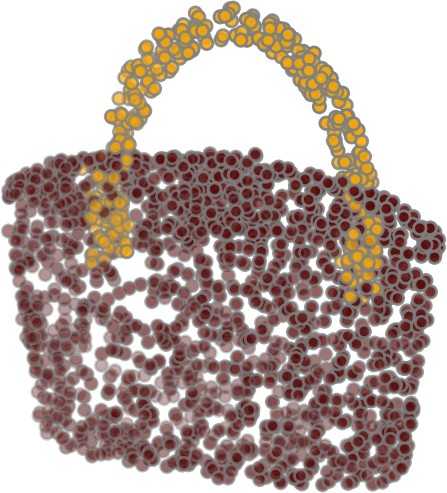} & \includegraphics[width=0.09\textwidth,height=1.1cm]{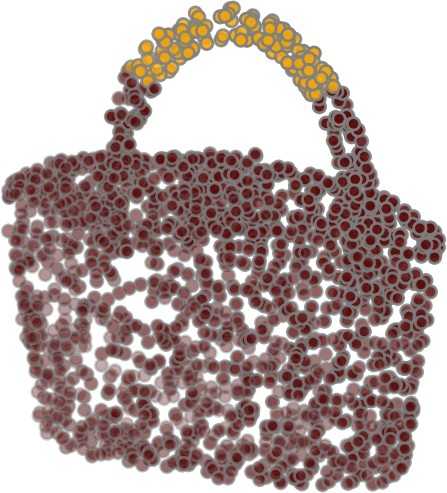} & \includegraphics[width=0.09\textwidth,height=1.1cm]{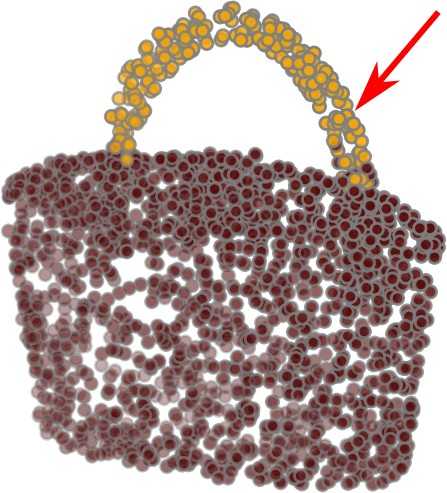} \\
		& 42.3\% & 96.8\% & & 69.6\% & 83.1\% & & 59.3\% & 72.1\% \\
		\midrule
		\includegraphics[width=0.09\textwidth,height=1.1cm]{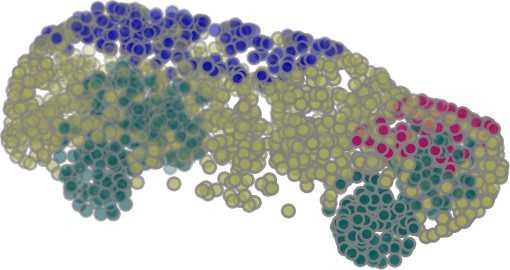} & \includegraphics[width=0.09\textwidth,height=1.1cm]{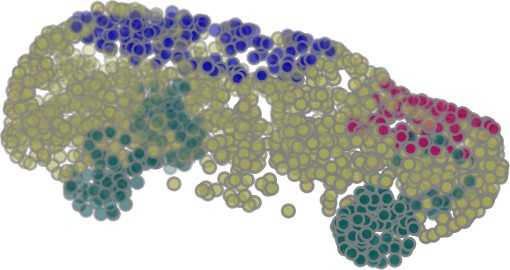} & \includegraphics[width=0.09\textwidth,height=1.1cm]{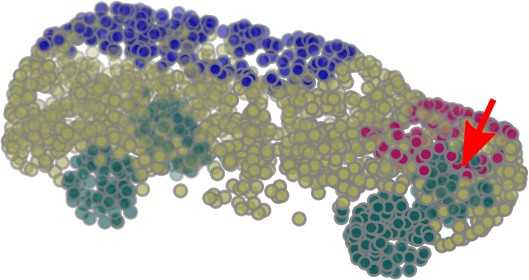} & \includegraphics[width=0.09\textwidth,height=1.1cm]{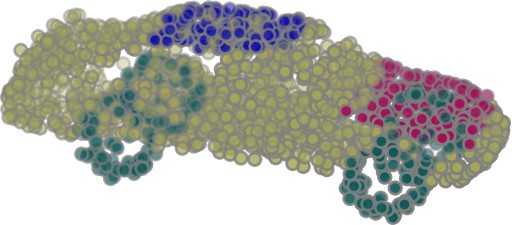} & \includegraphics[width=0.09\textwidth,height=1.1cm]{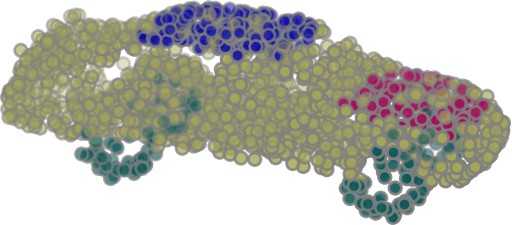}& \includegraphics[width=0.09\textwidth,height=1.1cm]{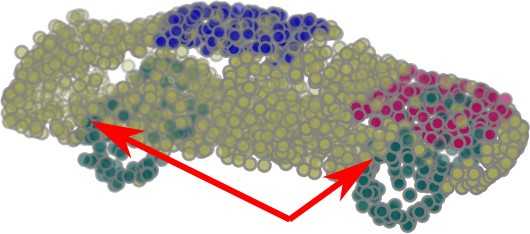} & \includegraphics[width=0.09\textwidth,height=1.1cm]{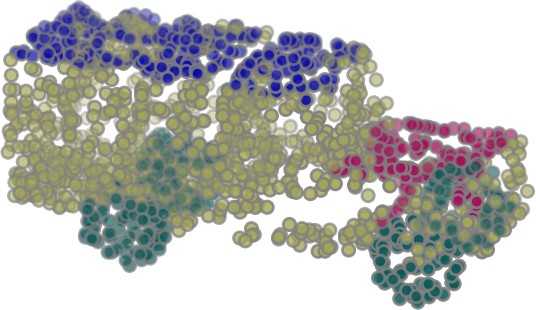} & \includegraphics[width=0.09\textwidth,height=1.1cm]{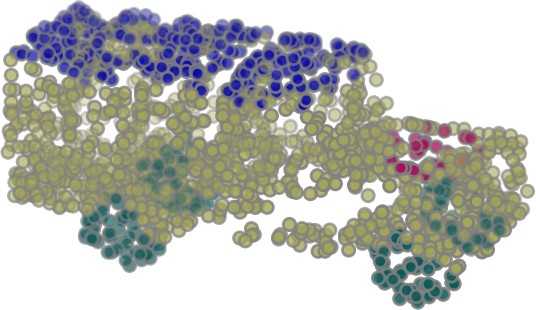} & \includegraphics[width=0.09\textwidth,height=1.1cm]{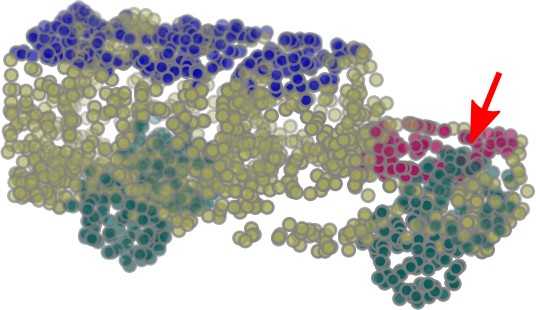} \\
		& 68.5\% & 82.3\% & & 70.8\% & 83.8\% & & 61.4\% & 79.0\% \\
		\midrule
		\includegraphics[width=0.09\textwidth,height=1.1cm]{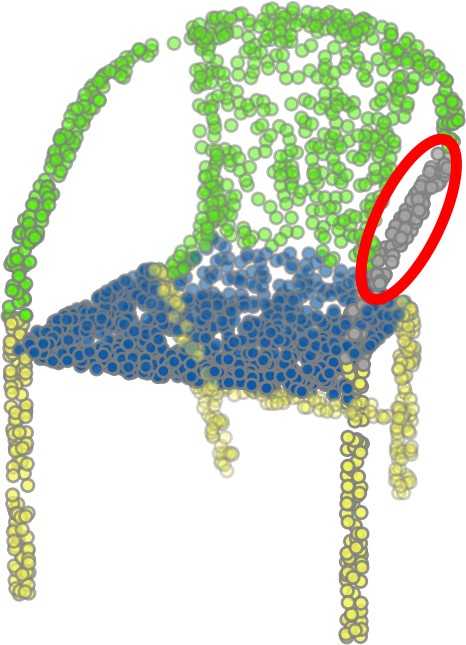} & \includegraphics[width=0.09\textwidth,height=1.1cm]{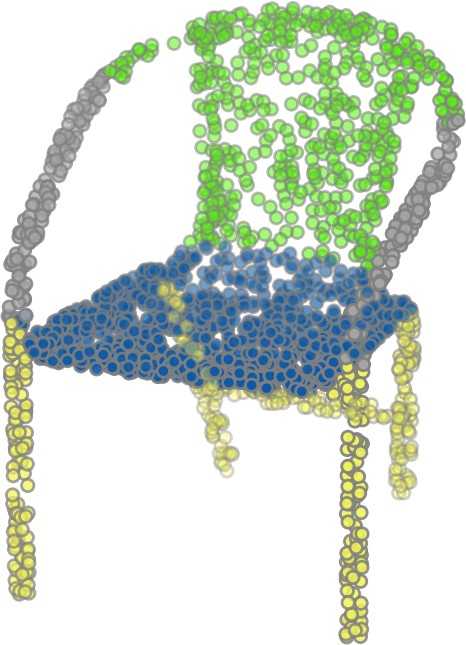} & \includegraphics[width=0.09\textwidth,height=1.1cm]{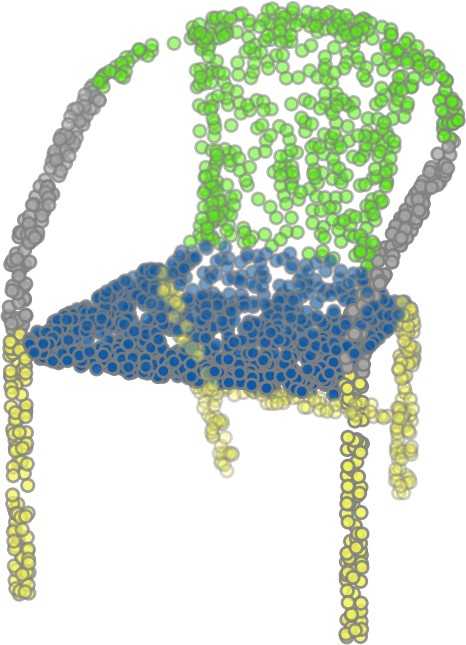} & \includegraphics[width=0.09\textwidth,height=1.1cm]{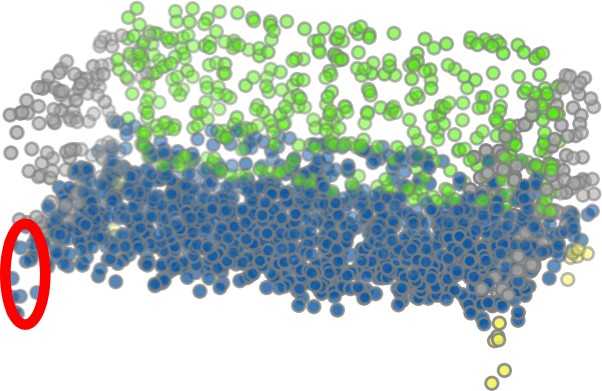} & \includegraphics[width=0.09\textwidth,height=1.1cm]{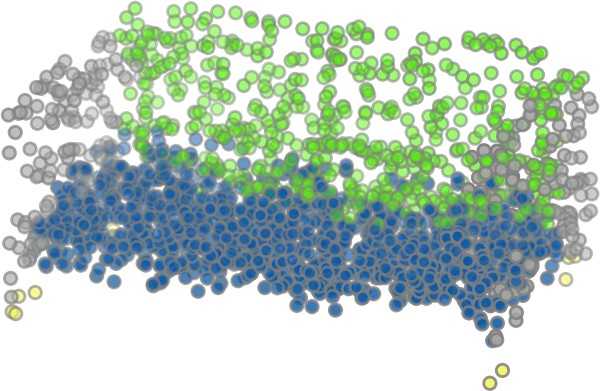}& \includegraphics[width=0.09\textwidth,height=1.1cm]{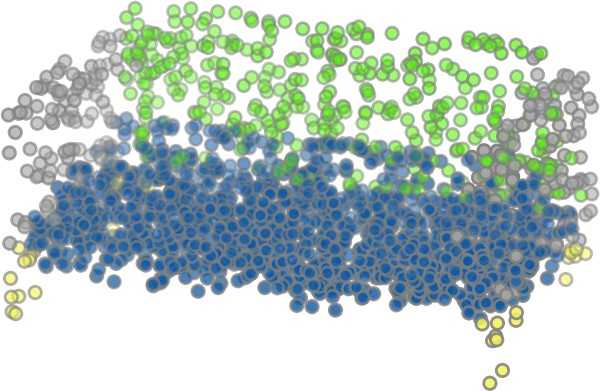} & \includegraphics[width=0.09\textwidth,height=1.1cm]{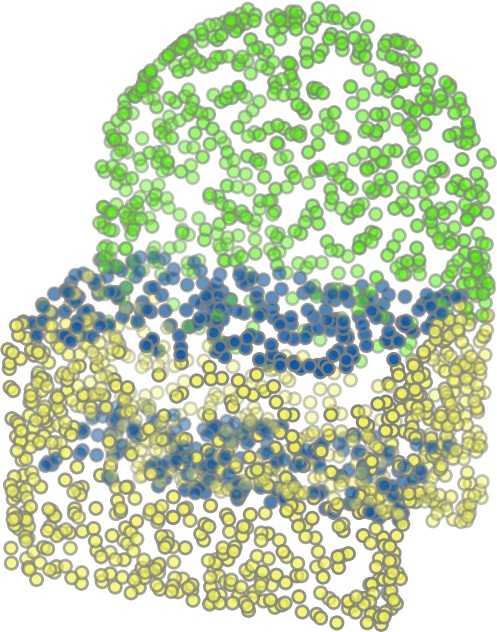} & \includegraphics[width=0.09\textwidth,height=1.1cm]{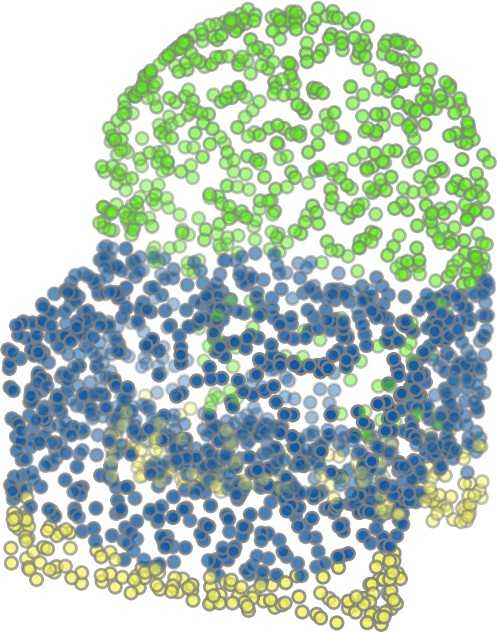} & \includegraphics[width=0.09\textwidth,height=1.1cm]{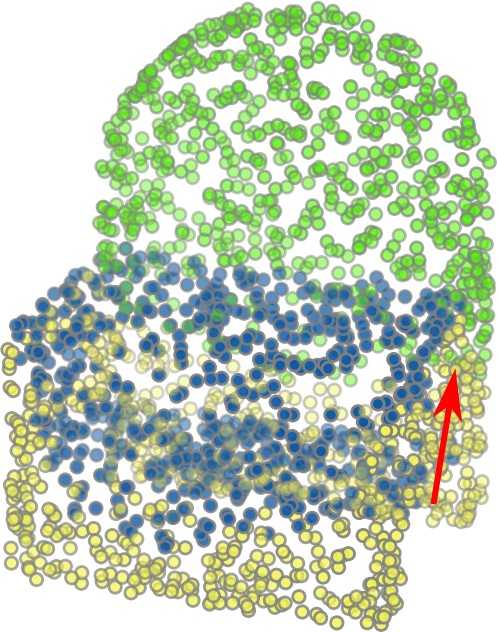} \\
		& 76.5\% & 78.3\% & & 59.5\% & 66.9\% & & 63.5\% & 82.8\% \\
		\midrule
		\includegraphics[width=0.09\textwidth,height=1.1cm]{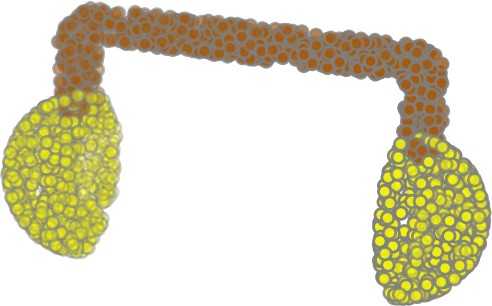} & \includegraphics[width=0.09\textwidth,height=1.1cm]{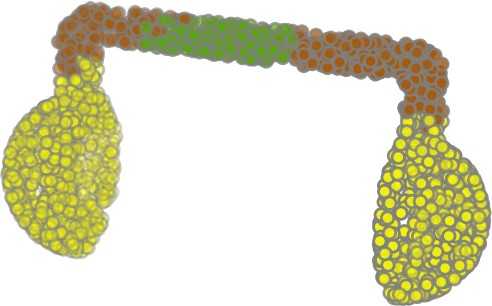} & \includegraphics[width=0.09\textwidth,height=1.1cm]{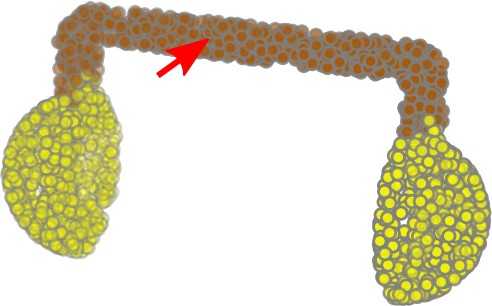} & \includegraphics[width=0.09\textwidth,height=1.1cm]{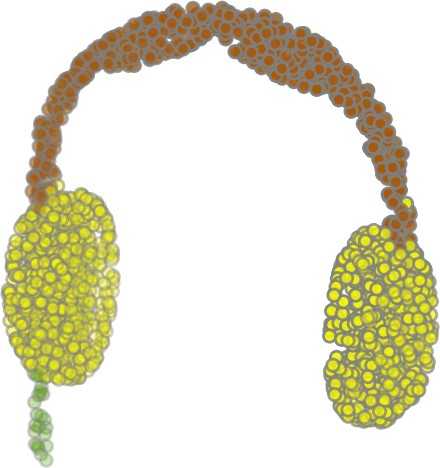} & \includegraphics[width=0.09\textwidth,height=1.1cm]{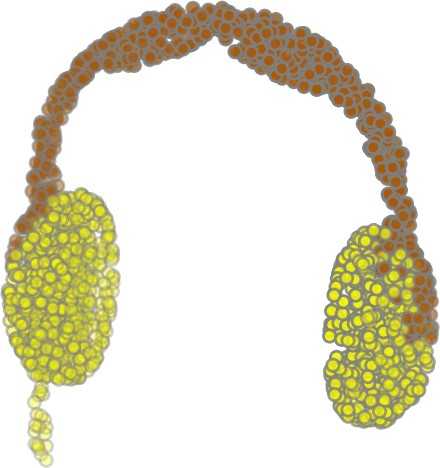}& \includegraphics[width=0.09\textwidth,height=1.1cm]{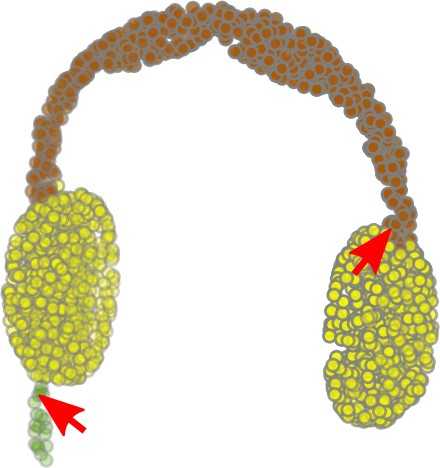} & \includegraphics[width=0.09\textwidth,height=1.1cm]{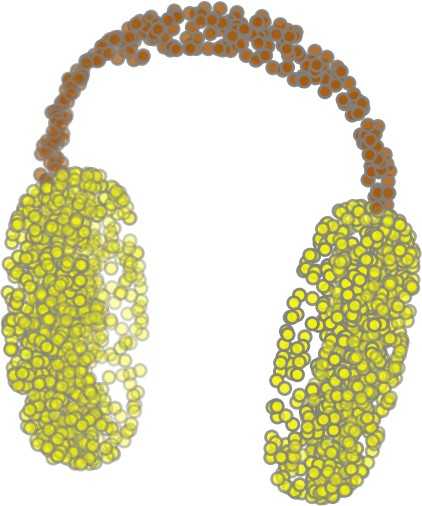} & \includegraphics[width=0.09\textwidth,height=1.1cm]{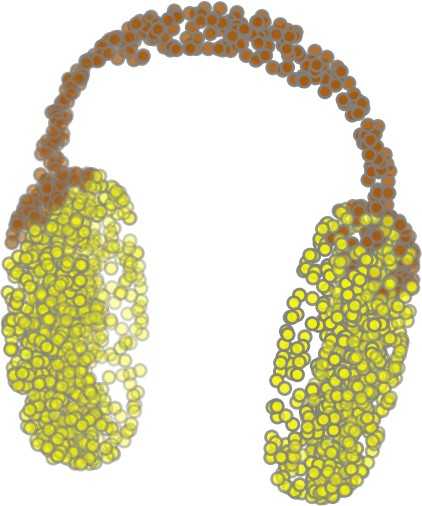} & \includegraphics[width=0.09\textwidth,height=1.1cm]{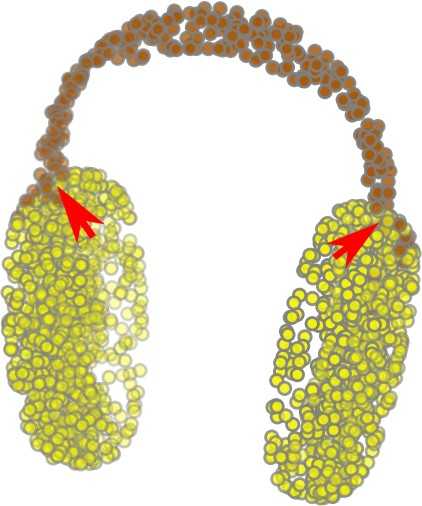} \\
		& 48.4\% & 93.6\% & & 55.8\% & 90.6\% & & 84.6\% & 94.9\% \\
		\midrule
		\includegraphics[width=0.07\textwidth,height=1.1cm]{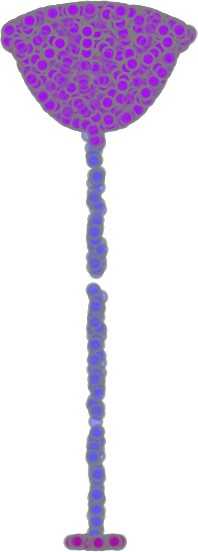} & \includegraphics[width=0.07\textwidth,height=1.1cm]{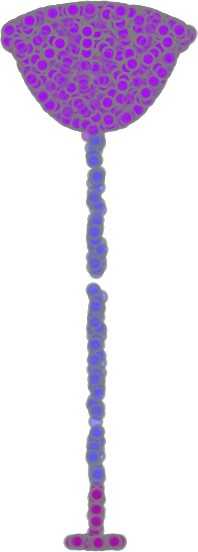} & \includegraphics[width=0.07\textwidth,height=1.1cm]{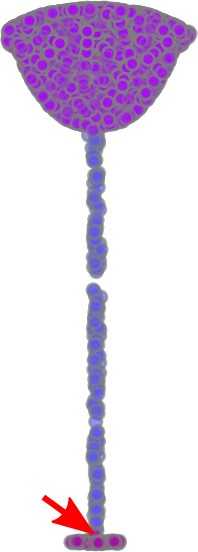} & \includegraphics[width=0.07\textwidth,height=1.1cm]{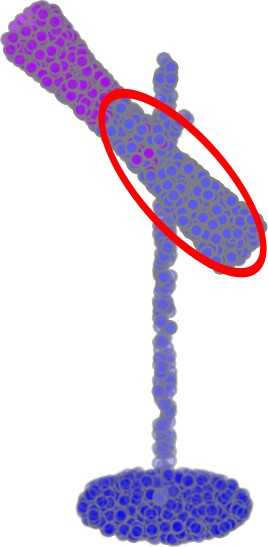} & \includegraphics[width=0.07\textwidth,height=1.1cm]{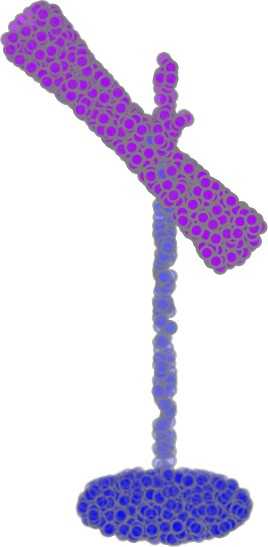}& \includegraphics[width=0.07\textwidth,height=1.1cm]{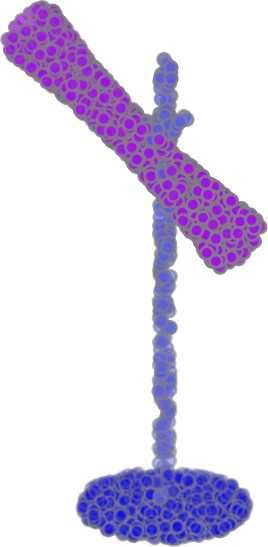} & \includegraphics[width=0.07\textwidth,height=1.1cm]{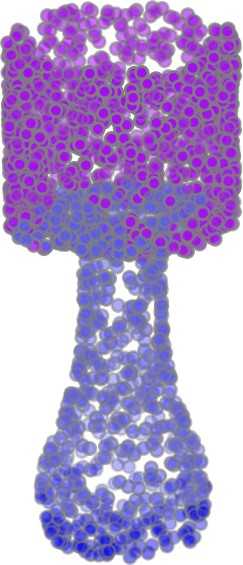} & \includegraphics[width=0.07\textwidth,height=1.1cm]{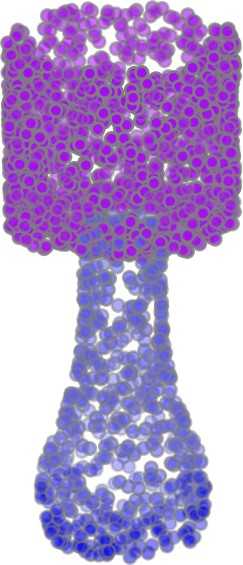} & \includegraphics[width=0.07\textwidth,height=1.1cm]{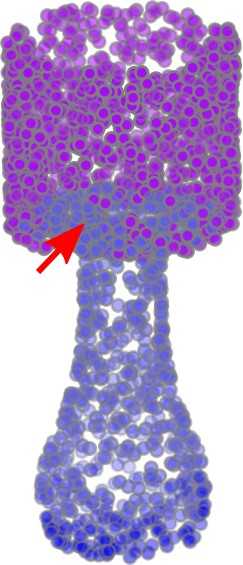} \\
		& 88.8\% & 96.8\% & & 64.0\% & 67.9\% & & 84.1\% & 91.5\% \\
		\midrule
		\includegraphics[width=0.09\textwidth,height=1.1cm]{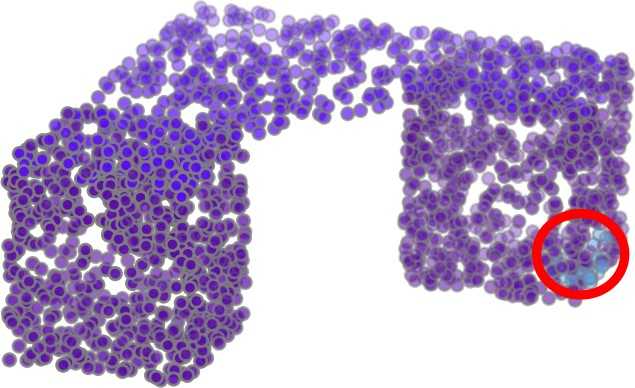} & \includegraphics[width=0.09\textwidth,height=1.1cm]{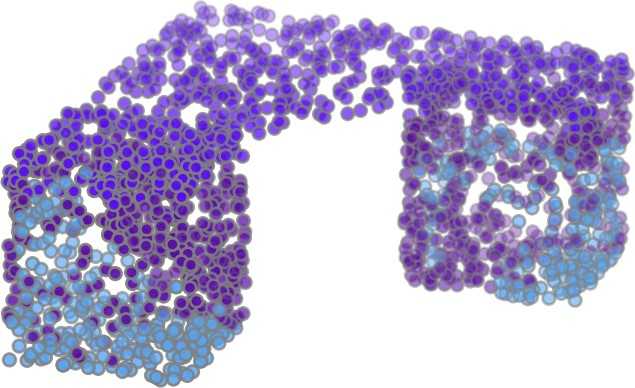} & \includegraphics[width=0.09\textwidth,height=1.1cm]{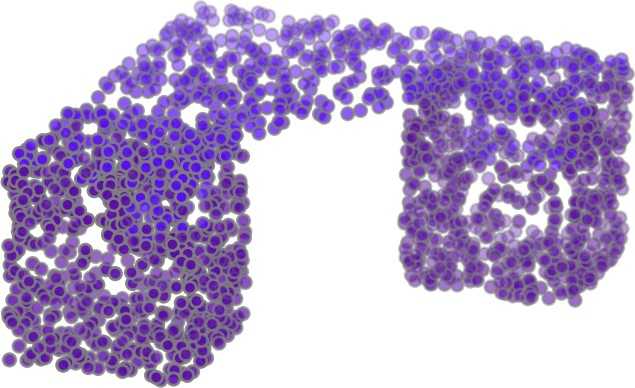} & \includegraphics[width=0.09\textwidth,height=1.1cm]{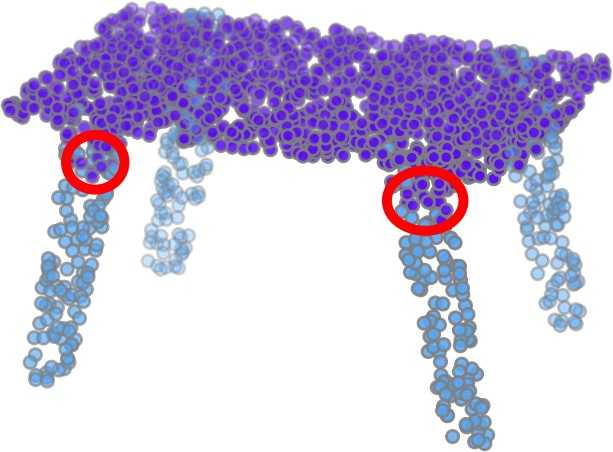} & \includegraphics[width=0.09\textwidth,height=1.1cm]{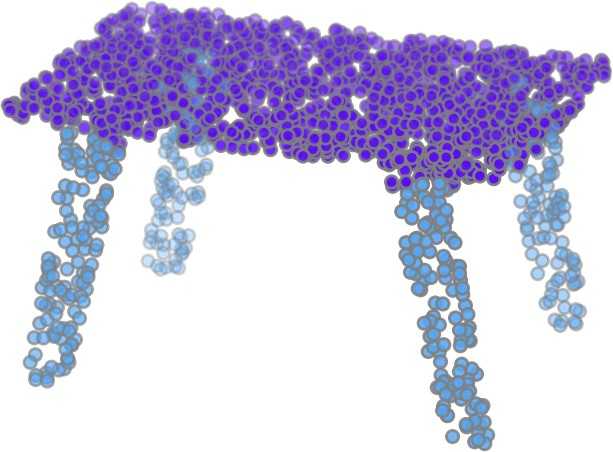}& \includegraphics[width=0.09\textwidth,height=1.1cm]{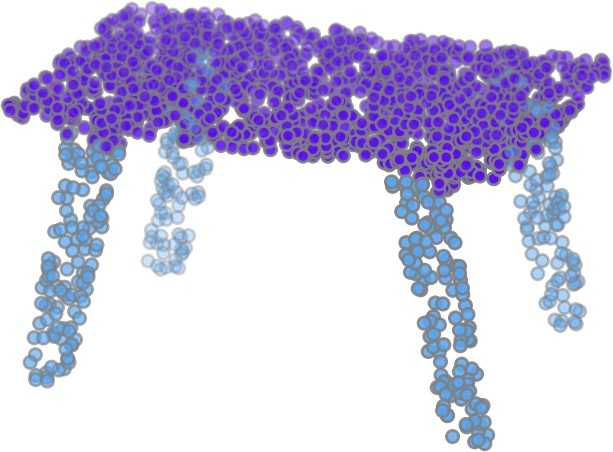} & \includegraphics[width=0.09\textwidth,height=1.1cm]{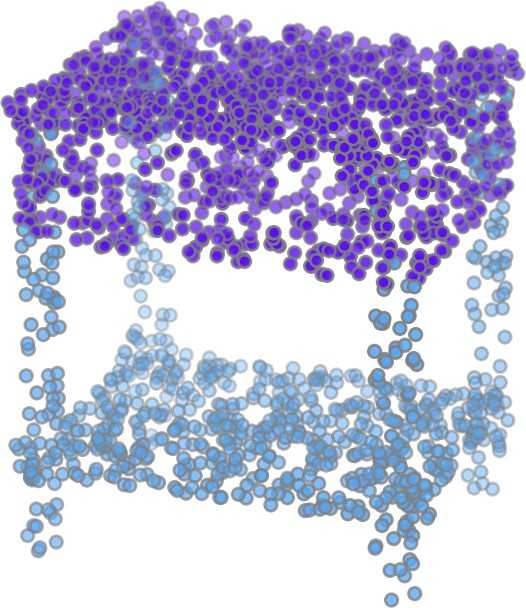} & \includegraphics[width=0.09\textwidth,height=1.1cm]{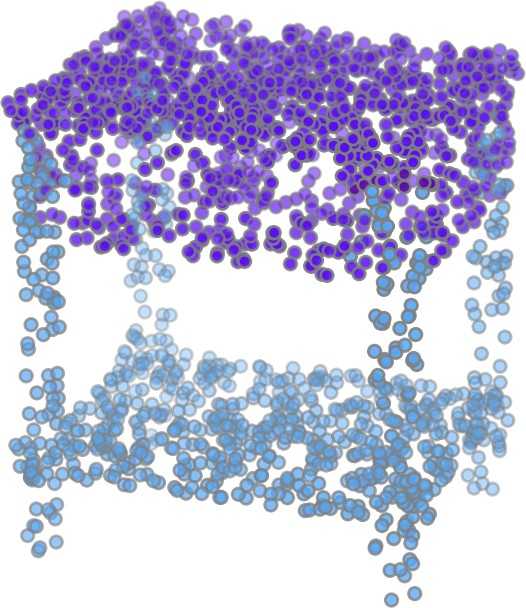} & \includegraphics[width=0.09\textwidth,height=1.1cm]{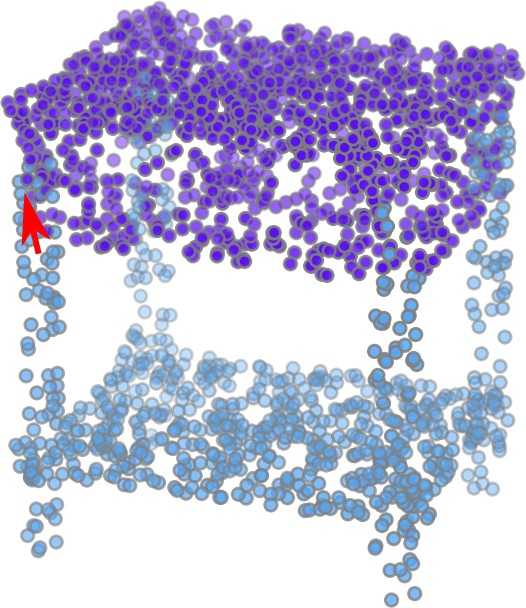} \\
		& 48.7\% & 57.9\% & & 89.9\% & 92.8\% & & 59.2\% & 93.4\% \\
		\midrule
		\includegraphics[width=0.09\textwidth,height=1.1cm]{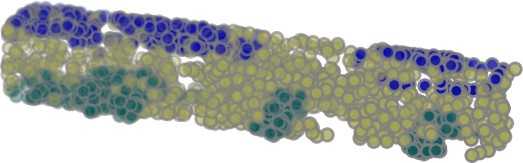} & \includegraphics[width=0.09\textwidth,height=1.1cm]{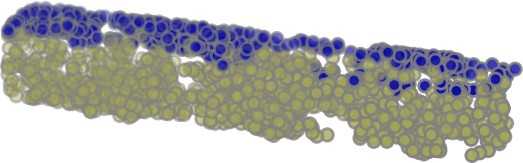} & \includegraphics[width=0.09\textwidth,height=1.1cm]{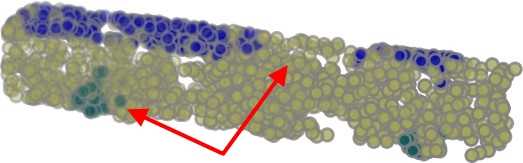} & \includegraphics[width=0.09\textwidth,height=1.1cm]{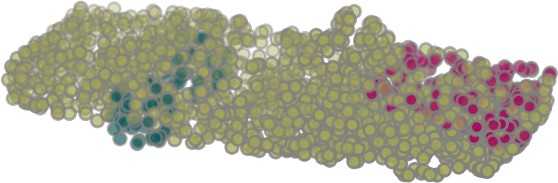} & \includegraphics[width=0.09\textwidth,height=1.1cm]{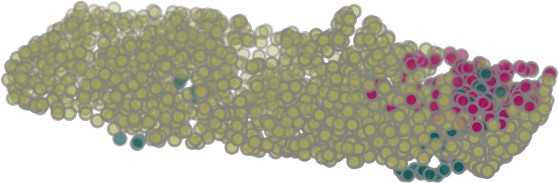}& \includegraphics[width=0.09\textwidth,height=1.1cm]{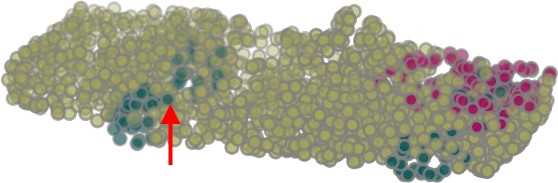} & \includegraphics[width=0.09\textwidth,height=1.1cm]{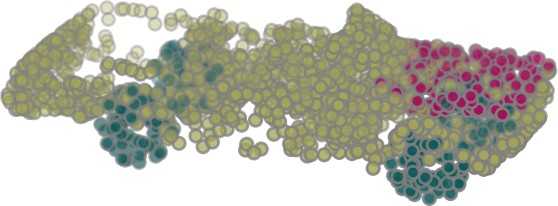} & \includegraphics[width=0.09\textwidth,height=1.1cm]{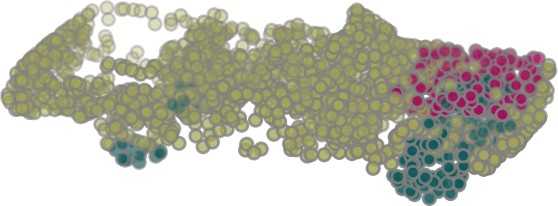} & \includegraphics[width=0.09\textwidth,height=1.1cm]{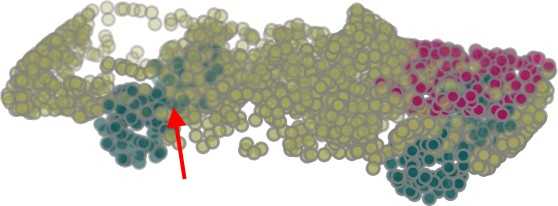} \\
		& 58.2\% & 65.4\% & & 63.2\% & 68.8\% & & 82.7\% & 90.4\% \\
		\midrule
		\includegraphics[width=0.09\textwidth,height=1.1cm]{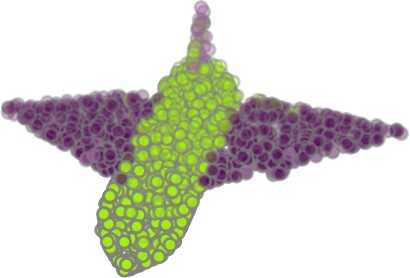} & \includegraphics[width=0.09\textwidth,height=1.1cm]{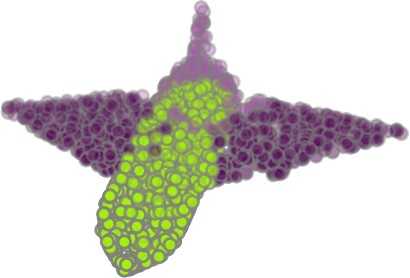} & \includegraphics[width=0.09\textwidth,height=1.1cm]{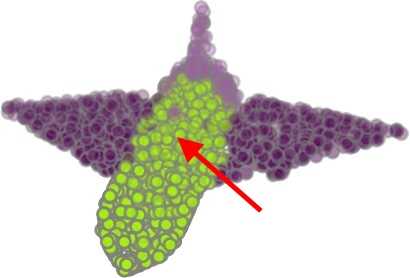} & \includegraphics[width=0.09\textwidth,height=1.1cm]{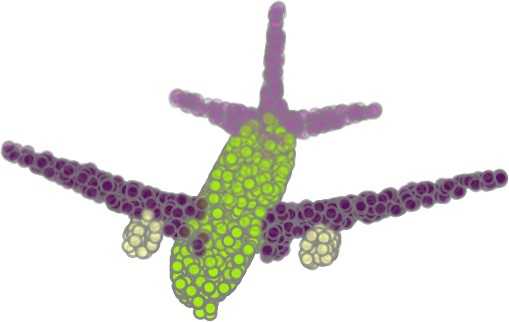} & \includegraphics[width=0.09\textwidth,height=1.1cm]{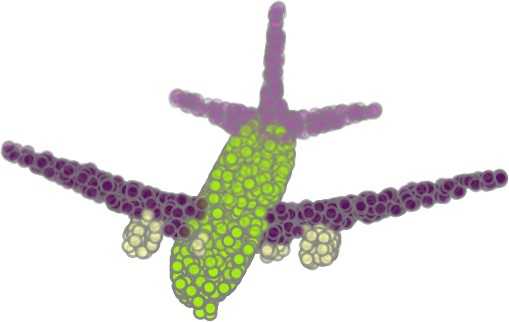}& \includegraphics[width=0.09\textwidth,height=1.1cm]{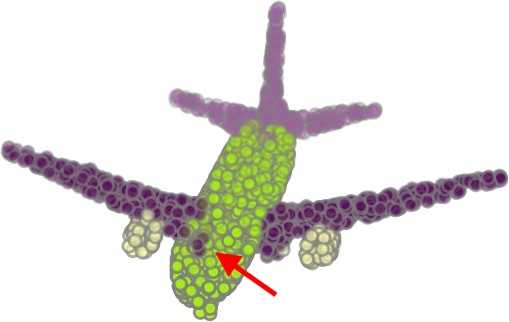} & \includegraphics[width=0.09\textwidth,height=1.1cm]{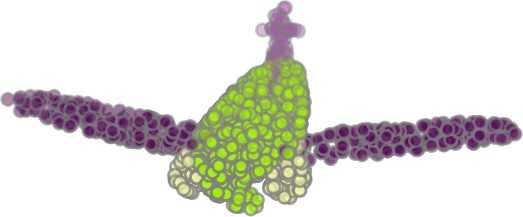} & \includegraphics[width=0.09\textwidth,height=1.1cm]{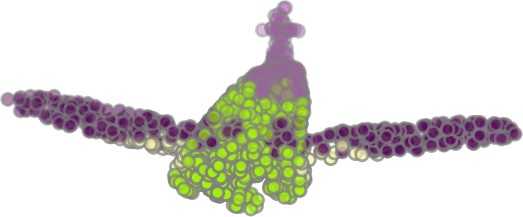} & \includegraphics[width=0.09\textwidth,height=1.1cm]{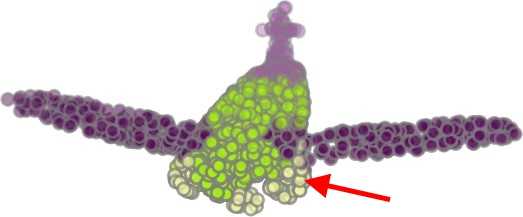} \\
		& 68.5\% & 74.0\% & & 90.8\% & 93.2\% & & 39.8\% & 59.1\% \\
		\midrule
		\includegraphics[width=0.09\textwidth,height=1.1cm]{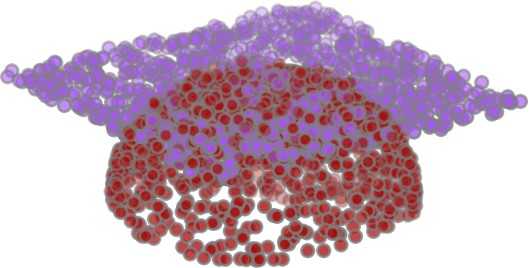} & \includegraphics[width=0.09\textwidth,height=1.1cm]{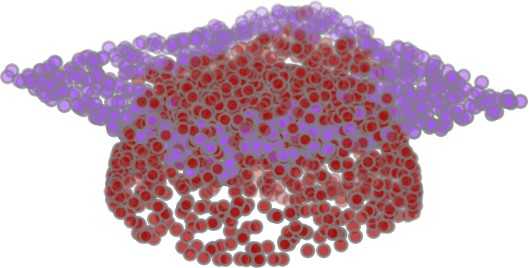} & \includegraphics[width=0.09\textwidth,height=1.1cm]{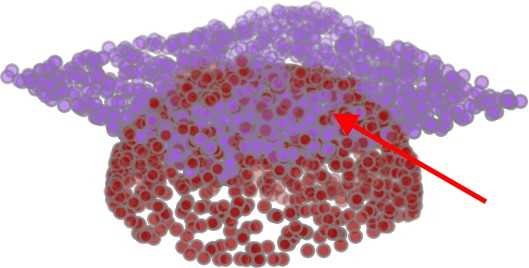} & \includegraphics[width=0.09\textwidth,height=1.1cm]{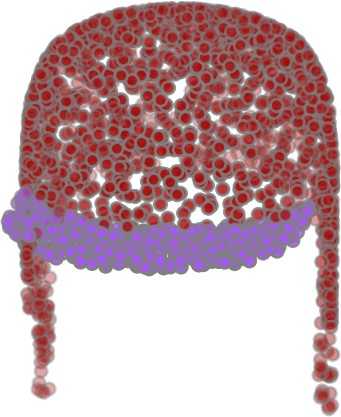} & \includegraphics[width=0.09\textwidth,height=1.1cm]{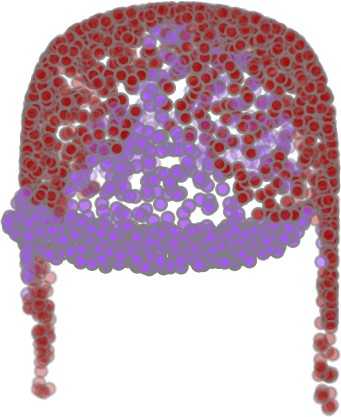}& \includegraphics[width=0.09\textwidth,height=1.1cm]{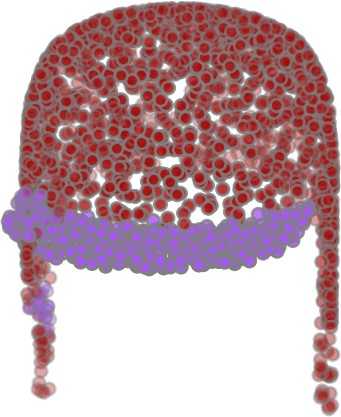} & \includegraphics[width=0.09\textwidth,height=1.1cm]{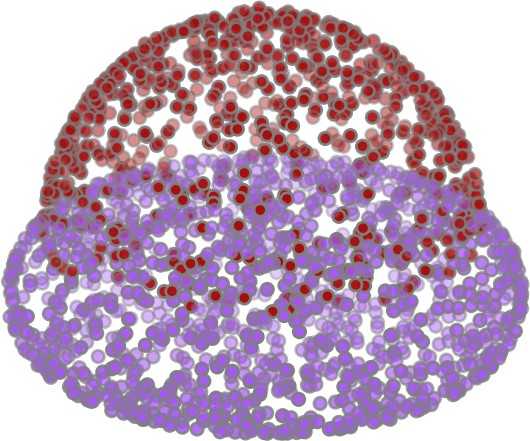} & \includegraphics[width=0.09\textwidth,height=1.1cm]{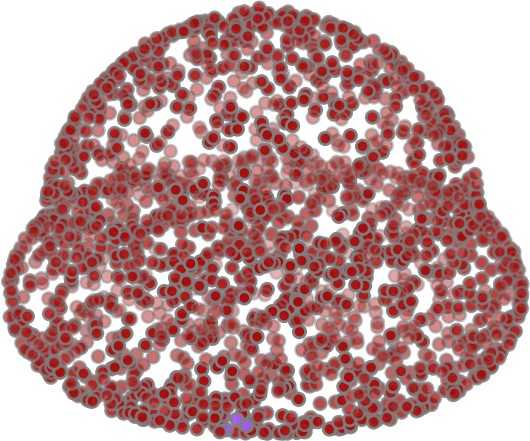} & \includegraphics[width=0.09\textwidth,height=1.1cm]{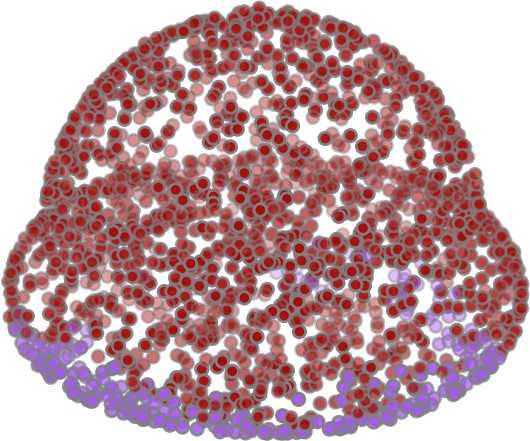} \\
		& 77.2\% & 91.6\% & & 69.4\% & 94.3\% & & 40.2\% & 53.9\% \\
		\bottomrule  
	\end{tabular} \vspace{-3mm}
	\caption{Examples of part segmentation results on ShapeNet part test dataset.
		IoU (\%) is listed below each result for reference.
		Red arrows: KCNet improvements.
		Red circles: some errors in ground truth (GT).
		Better viewed in color.
		\label{fig:segmentation-comp}
	}
	\vspace{-2mm}
\end{figure*}

\noindent\\
\textbf{Network Configuration.} Segmentation network has 10 parametric layers. 
Features of different layers capturing local features are concatenated with the replicated global features and shape information, as in~\cite{qi2017pointnet}.
Details are in Figure~\ref{fig:arch-segmentation}.
Again, ReLU is used in each layer without Batchnorm. 
Dropout layers are used for fully connected layers with drop ratio 0.3.
We used 18-NN graph for kernel computation and graph max pooling.
$L=16$ sets of kernels are used, in which each kernel has $M=18$ points uniformly initialized within $[-0.2, 0.2]$ and kernel width $\sigma=0.005$.
Other hyper-parameters are the same as in shape classification.
No data augmentation was performed.

\noindent\\
\textbf{Results.} We compared our method with PointNet~\cite{qi2017pointnet}, PointNet++~\cite{qi2017pointnetplusplus} and Kd-Net~\cite{klokov2017escape}. 
We use intersection over union (IoU) of each category as the evaluation metric following~\cite{klokov2017escape}: IoU of each shape instance is the average IoU of each part that occurs in this shape category (the IoUs of the parts belonging to other shape categories are ignored following the protocol of \cite{qi2017pointnet}). 
The mean IoU (mIoU) of each category is obtained by averaging IoUs of all the shapes in that category. 
The overall average instance mIoU (Ins. mIoU) is calculated by averaging IoUs of all the shape instances. 
Besides, we also report overall average category mIoU (Cat. mIoU) that is directly averaged over 16 categories.
Table~\ref{table:part-segmentation} lists the results.
Compared with PointNet++ that uses surface normals as additional inputs, our KCNet only takes raw point clouds as input and achieves better performance with more efficiency regarding computation and parameters in Table~\ref{table:time-model-size}.
Figure~\ref{fig:segmentation-comp} displays some examples of predicted results on ShapeNet part test dataset.

\subsection{Ablation Study}
\label{experiment:analysis}
In this section, we further conducted several ablation experiments, investigating various design variations and demonstrating the advantages of KCNet.

\begin{table}[!t]
	\centering
	\begin{tabular}{l | c} \hline 
		\textbf{Effectiveness of Kernel Correlation} & Accuracy (\%)\\ \hline
		Normal & 88.4 \\
		Kernel correlation & \textbf{90.5} \\ \hline \hline
		\textbf{Symmetric Functions} & Accuracy (\%)\\ \hline 
		Graph average pooling & 88.0 \\ 
		Graph max pooling & \textbf{88.6} \\ \hline \hline
		\textbf{Effectiveness of Local Structures} & Accuracy (\%)\\ \hline
		Baseline: PointNet (vanilla) & 87.2 \\
		Kernel correlation (geometric) & 90.5 \\
		Graph max pooling (feature) & 88.6 \\
		Both & \textbf{91.0} \\ \hline
	\end{tabular} \vspace{-1mm}
	\caption{Ablation study on ModelNet40 test set.\label{table:model-analysis}}
\end{table}

\noindent\\
\textbf{Effectiveness of Kernel Correlation.} Table~\ref{table:model-analysis} lists the comparison between kernel correlation and normal. 
In this experiment, we used normals as local geometric features, concatenated them with coordinates and passed them into the proposed architecture in Figure~\ref{fig:arch-classification}. 
Normal of each point was computed by applying PCA to the covariance matrix to obtain the direction of minimal variance. Results show that kernel correlation is better than normals. 

\noindent\\
\textbf{Symmetric Functions.} Symmetric function is able to make a network invariant to input permutation ~\cite{qi2017pointnet}. 
In this experiment, we investigated the performance of graph max pooling and graph average pooling. 
As shown in Table \ref{table:model-analysis}, graph max pooling has a marginal improvement over graph average pooling, and is faster to compute, thus was adopted.

\noindent\\
\textbf{Effectiveness of Local Structures.}
In Table \ref{table:model-analysis} we also demonstrate the effect of our local geometric and feature structures learned by kernel correlation and graph max pooling, respectively. 
Note that our kernel correlation and graph max pooling layer already respectively achieve comparable or even better performances compared to PointNet.

\begin{table}[!t]
	\centering
	\begin{tabular}{lc|lc|lc} \hline 
		$L$ & Acc. (\%) & $M$ & Acc. (\%) & $\sigma$ & Acc. (\%) \\ \hline
		16  & 90.7      & 3   & 90.9      & $1\mathrm{e}{-3}$ & 90.0 \\ \hline
\textbf{32} & \textbf{91.0}      & 8   & 90.4      & $\mathbf{5\mathrm{\mathbf{e}}{-3}}$ & \textbf{91.0}  \\ \hline
		48  & 91.0  & \textbf{16}  & \textbf{91.0}      & $1\mathrm{e}{-2}$ & 90.4  \\ \hline
	\end{tabular}
	\caption{
		Choosing hyper-parameters.
		Each column only changes the corresponding parameter (base setting in bold).\label{table:hyper-parameters}
	} \vspace{-6mm}
\end{table}

\noindent\\
\textbf{Choosing Hyper-parameters.} KCNet have several unique hyper-parameters: $L$, $M$, and $\sigma$, as explained in Section~\ref{method:KC}. We report their individual influences in Table~\ref{table:hyper-parameters}.

\begin{figure}[t]
	\centering
	\includegraphics[width=0.88\columnwidth]{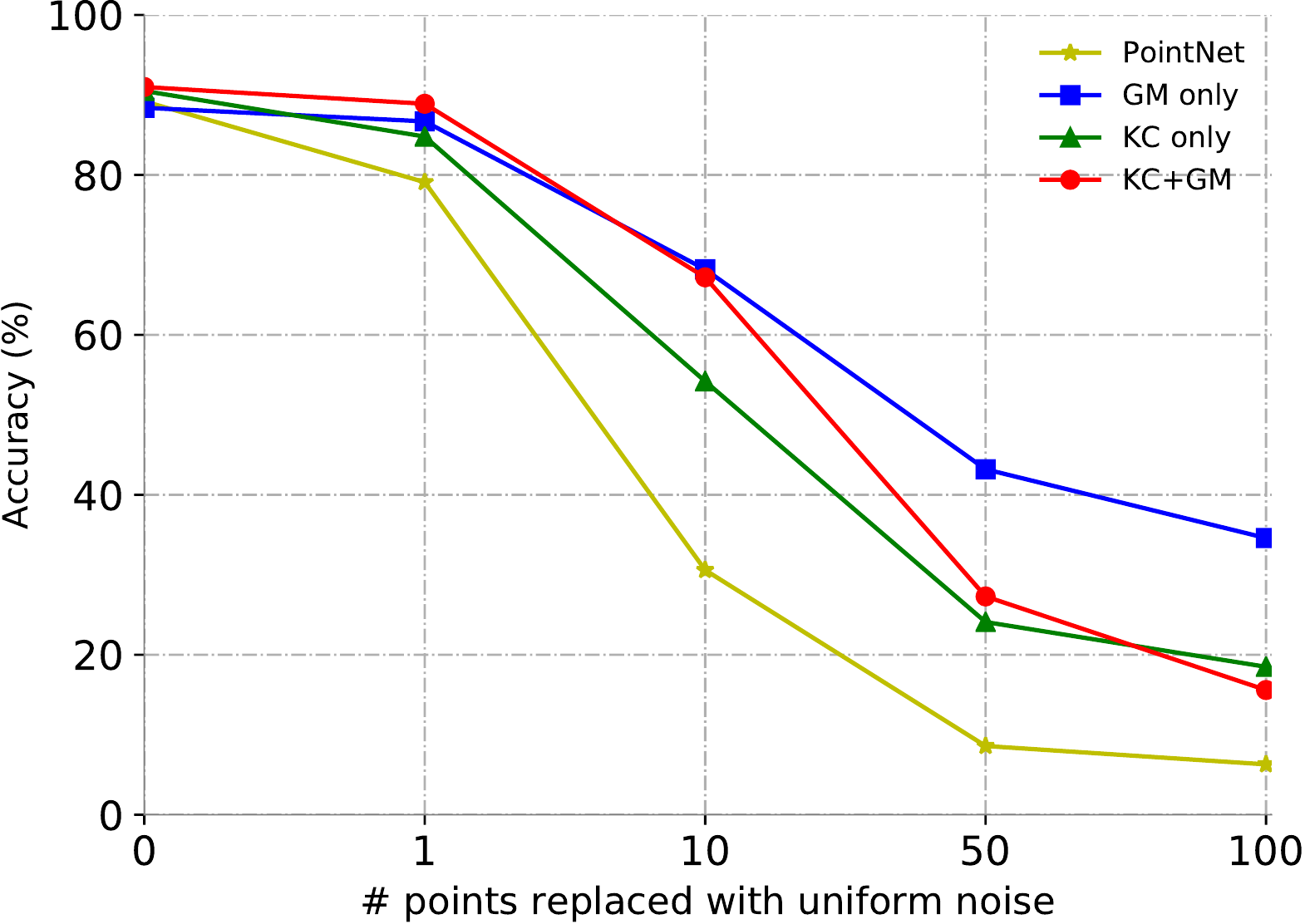}
	\caption{KCNet vs. PointNet on random noise. Different numbers of points in each object are replaced with uniform noise between [-1,1]. Metric is overall classification accuracy on the ModelNet40 test set. KCNet is more robust against random noise. GM only: graph max pooling only. KC only: kernel correlation only. KC+GM: both.\label{fig:random-noise}}
	\vspace{-2mm}
\end{figure}

\noindent\\
\textbf{Robustness Test.} 
\label{robutness-test}
We compared our networks with PointNet on robustness against random noise in the input point cloud. Both networks are trained on the same train and test data with 1024 points per object. The PointNet was trained with the same data augmentation in~\cite{qi2017pointnet} using the authors' code. Our networks were trained without data augmentation.  During testing, a certain number of randomly selected input points were replaced with uniformly distributed noise ranging  [-1.0, 1.0]. As shown in Figure~\ref{fig:random-noise}, our networks are more robust against random noise. The accuracy of PointNet drops 58.6\% when 10 points were replaced with random noise (from 89.2\% to 30.6\%), while ours (KC+GM) only drops 23.8\% (from 91.0\% to 67.2\%).
Besides, it can be seen that within the groups of experiments, graph max pooling is the most robust under random noise.  
We speculate that it is caused by the local max pooling - neighbor points sharing max features along each dimension so random noise in the neighborhood could not easily affect the prediction.
This could also explain why KC+GM is more robust than KC only.
This test shows an advantage of local structures over per-point features - our network learns to exploit local geometric and feature structures within neighboring regions and thus is robust against random noise.

\section{Conclusion}
\label{conclusion}
We proposed kernel correlation and graph pooling to improve PointNet-like methods.
Experiments have shown that our method efficiently captures local patterns and robustly improves performances of 3D point cloud semantic learning.
We will generalize the kernel correlation to higher dimensions, with learnable kernel widths in the future.

\section*{Acknowledgment}

The authors gratefully acknowledge the helpful comments and suggestions from Teng-Yok Lee, Ziming Zhang, Zhiding Yu, Yuichi Taguchi, and Alan Sullivan.

\cleardoublepage
\newpage

{\small
\bibliographystyle{ieee}
\bibliography{references}

\begin{thebibliography}{10}\itemsep=-1pt

\bibitem{atwood2016diffusion}
J.~Atwood and D.~Towsley.
\newblock Diffusion-convolutional neural networks.
\newblock In {\em Advances in Neural Information Processing Systems}, pages
  1993--2001, 2016.

\bibitem{brock2016generative}
A.~Brock, T.~Lim, J.~M. Ritchie, and N.~Weston.
\newblock Generative and discriminative voxel modeling with convolutional
  neural networks.
\newblock {\em arXiv preprint arXiv:1608.04236}, 2016.

\bibitem{bronstein2017geometric}
M.~M. Bronstein, J.~Bruna, Y.~LeCun, A.~Szlam, and P.~Vandergheynst.
\newblock Geometric deep learning: going beyond euclidean data.
\newblock {\em IEEE Signal Processing Magazine}, 34(4):18--42, 2017.

\bibitem{bruna2013spectral}
J.~Bruna, W.~Zaremba, A.~Szlam, and Y.~LeCun.
\newblock Spectral networks and locally connected networks on graphs.
\newblock {\em arXiv preprint arXiv:1312.6203}, 2013.

\bibitem{chen1991object}
Y.~Chen and G.~Medioni.
\newblock Object modeling by registration of multiple range images.
\newblock In {\em IEEE International Conference on Robotics and Automation
  (ICRA)}, pages 2724--2729, 1991.

\bibitem{cignoni2008meshlab}
P.~Cignoni, M.~Callieri, M.~Corsini, M.~Dellepiane, F.~Ganovelli, and
  G.~Ranzuglia.
\newblock {MeshLab: an Open-Source Mesh Processing Tool}.
\newblock In V.~Scarano, R.~D. Chiara, and U.~Erra, editors, {\em Eurographics
  Italian Chapter Conference}. The Eurographics Association, 2008.

\bibitem{dai2017scannet}
A.~Dai, A.~X. Chang, M.~Savva, M.~Halber, T.~Funkhouser, and M.~Nie{\ss}ner.
\newblock Scannet: Richly-annotated 3d reconstructions of indoor scenes.
\newblock In {\em Proc. IEEE Conf. on Computer Vision and Pattern Recognition
  (CVPR)}, volume~1, 2017.

\bibitem{defferrard2016convolutional}
M.~Defferrard, X.~Bresson, and P.~Vandergheynst.
\newblock Convolutional neural networks on graphs with fast localized spectral
  filtering.
\newblock In {\em Advances in Neural Information Processing Systems}, pages
  3844--3852, 2016.

\bibitem{duvenaud2015convolutional}
D.~K. Duvenaud, D.~Maclaurin, J.~Iparraguirre, R.~Bombarell, T.~Hirzel,
  A.~Aspuru-Guzik, and R.~P. Adams.
\newblock Convolutional networks on graphs for learning molecular fingerprints.
\newblock In {\em Advances in neural information processing systems}, pages
  2224--2232, 2015.

\bibitem{edwards2016graph}
M.~Edwards and X.~Xie.
\newblock Graph based convolutional neural network.
\newblock {\em arXiv preprint arXiv:1609.08965}, 2016.

\bibitem{feng2014fast}
C.~Feng, Y.~Taguchi, and V.~R. Kamat.
\newblock Fast plane extraction in organized point clouds using agglomerative
  hierarchical clustering.
\newblock In {\em IEEE International Conference on Robotics and Automation
  (ICRA)}, pages 6218--6225, 2014.

\bibitem{golovinskiy2009shape}
A.~Golovinskiy, V.~G. Kim, and T.~Funkhouser.
\newblock Shape-based recognition of 3d point clouds in urban environments.
\newblock In {\em IEEE International Conference on Computer Vision (ICCV)},
  pages 2154--2161, 2009.

\bibitem{henaff2015deep}
M.~Henaff, J.~Bruna, and Y.~LeCun.
\newblock Deep convolutional networks on graph-structured data.
\newblock {\em arXiv preprint arXiv:1506.05163}, 2015.

\bibitem{holz2013fast}
D.~Holz and S.~Behnke.
\newblock Fast range image segmentation and smoothing using approximate surface
  reconstruction and region growing.
\newblock {\em Intelligent autonomous systems 12}, pages 61--73, 2013.

\bibitem{hoppe1992surface}
H.~Hoppe, T.~DeRose, T.~Duchamp, J.~McDonald, and W.~Stuetzle.
\newblock Surface reconstruction from unorganized points.
\newblock {\em SIGGRAPH Comput. Graph.}, 26(2):71--78, July 1992.

\bibitem{jia2014caffe}
Y.~Jia, E.~Shelhamer, J.~Donahue, S.~Karayev, J.~Long, R.~Girshick,
  S.~Guadarrama, and T.~Darrell.
\newblock Caffe: Convolutional architecture for fast feature embedding.
\newblock In {\em Proceedings of the 22nd ACM international conference on
  Multimedia}, pages 675--678, 2014.

\bibitem{jian2011robust}
B.~Jian and B.~C. Vemuri.
\newblock Robust point set registration using gaussian mixture models.
\newblock {\em IEEE Transactions on Pattern Analysis and Machine Intelligence},
  33(8):1633--1645, 2011.

\bibitem{kempe2003maximizing}
D.~Kempe, J.~Kleinberg, and {\'E}.~Tardos.
\newblock Maximizing the spread of influence through a social network.
\newblock In {\em Proceedings of the ninth ACM SIGKDD international conference
  on Knowledge discovery and data mining}, pages 137--146. ACM, 2003.

\bibitem{kipf2016semi}
T.~N. Kipf and M.~Welling.
\newblock Semi-supervised classification with graph convolutional networks.
\newblock {\em arXiv preprint arXiv:1609.02907}, 2016.

\bibitem{klokov2017escape}
R.~Klokov and V.~Lempitsky.
\newblock Escape from cells: Deep kd-networks for the recognition of 3d point
  cloud models.
\newblock {\em International Conference on Computer Vision (ICCV)}, 2017.

\bibitem{lecun1998gradient}
Y.~LeCun, L.~Bottou, Y.~Bengio, and P.~Haffner.
\newblock Gradient-based learning applied to document recognition.
\newblock {\em Proceedings of the IEEE}, 86(11):2278--2324, 1998.

\bibitem{levie2017cayleynets}
R.~Levie, F.~Monti, X.~Bresson, and M.~M. Bronstein.
\newblock Cayleynets: Graph convolutional neural networks with complex rational
  spectral filters.
\newblock {\em arXiv preprint arXiv:1705.07664}, 2017.

\bibitem{li2016fpnn}
Y.~Li, S.~Pirk, H.~Su, C.~R. Qi, and L.~J. Guibas.
\newblock Fpnn: Field probing neural networks for 3d data.
\newblock In {\em Advances in Neural Information Processing Systems}, pages
  307--315, 2016.

\bibitem{masci2015geodesic}
J.~Masci, D.~Boscaini, M.~Bronstein, and P.~Vandergheynst.
\newblock Geodesic convolutional neural networks on riemannian manifolds.
\newblock In {\em Proceedings of the IEEE international conference on computer
  vision workshops}, pages 37--45, 2015.

\bibitem{maturana2015voxnet}
D.~Maturana and S.~Scherer.
\newblock Voxnet: A 3d convolutional neural network for real-time object
  recognition.
\newblock In {\em IEEE International Conference on Intelligent Robots and
  Systems (IROS)}, pages 922--928. IEEE, 2015.

\bibitem{monti2016geometric}
F.~Monti, D.~Boscaini, J.~Masci, E.~Rodol{\`a}, J.~Svoboda, and M.~M.
  Bronstein.
\newblock Geometric deep learning on graphs and manifolds using mixture model
  cnns.
\newblock {\em arXiv preprint arXiv:1611.08402}, 2016.

\bibitem{niepert2016learning}
M.~Niepert, M.~Ahmed, and K.~Kutzkov.
\newblock Learning convolutional neural networks for graphs.
\newblock In {\em International conference on machine learning}, pages
  2014--2023, 2016.

\bibitem{ouyang2005normal}
D.~OuYang and H.-Y. Feng.
\newblock On the normal vector estimation for point cloud data from smooth
  surfaces.
\newblock {\em Computer-Aided Design}, 37(10):1071--1079, 2005.

\bibitem{qi2017pointnet}
C.~R. Qi, H.~Su, K.~Mo, and L.~J. Guibas.
\newblock Pointnet: Deep learning on point sets for 3d classification and
  segmentation.
\newblock {\em Proc. IEEE Conf. on Computer Vision and Pattern Recognition
  (CVPR)}, 2017.

\bibitem{qi2016volumetric}
C.~R. Qi, H.~Su, M.~Nie{\ss}ner, A.~Dai, M.~Yan, and L.~J. Guibas.
\newblock Volumetric and multi-view cnns for object classification on 3d data.
\newblock In {\em Proc. IEEE Conf. on Computer Vision and Pattern Recognition
  (CVPR)}, pages 5648--5656, 2016.

\bibitem{qi2017pointnetplusplus}
C.~R. Qi, L.~Yi, H.~Su, and L.~J. Guibas.
\newblock Pointnet++: Deep hierarchical feature learning on point sets in a
  metric space.
\newblock {\em Advances in Neural Information Processing Systems}, 2017.

\bibitem{riegler2017octnet}
G.~Riegler, A.~O. Ulusoy, and A.~Geiger.
\newblock Octnet: Learning deep 3d representations at high resolutions.
\newblock In {\em Proc. IEEE Conf. on Computer Vision and Pattern Recognition
  (CVPR)}, volume~3, 2017.

\bibitem{scott1991algorithm}
G.~L. Scott and H.~C. Longuet-Higgins.
\newblock An algorithm for associating the features of two images.
\newblock {\em Proceedings of the Royal Society of London B: Biological
  Sciences}, 244(1309):21--26, 1991.

\bibitem{simonovsky2017dynamic}
M.~Simonovsky and N.~Komodakis.
\newblock Dynamic edge-conditioned filters in convolutional neural networks on
  graphs.
\newblock {\em Proc. IEEE Conf. on Computer Vision and Pattern Recognition
  (CVPR)}, 2017.

\bibitem{strom2010graph}
J.~Strom, A.~Richardson, and E.~Olson.
\newblock Graph-based segmentation for colored 3d laser point clouds.
\newblock In {\em IEEE/RSJ International Conference on Intelligent Robots and
  Systems (IROS)}, pages 2131--2136, 2010.

\bibitem{su2015multi}
H.~Su, S.~Maji, E.~Kalogerakis, and E.~Learned-Miller.
\newblock Multi-view convolutional neural networks for 3d shape recognition.
\newblock In {\em International Conference on Computer Vision (ICCV)}, pages
  945--953, 2015.

\bibitem{thanou2016graph}
D.~Thanou, P.~A. Chou, and P.~Frossard.
\newblock Graph-based compression of dynamic 3d point cloud sequences.
\newblock {\em IEEE Transactions on Image Processing}, 25(4):1765--1778, 2016.

\bibitem{tsin2004correlation}
Y.~Tsin and T.~Kanade.
\newblock A correlation-based approach to robust point set registration.
\newblock In {\em European conference on computer vision (ECCV)}, pages
  558--569, 2004.

\bibitem{vosselman20013d}
G.~Vosselman, S.~Dijkman, et~al.
\newblock 3d building model reconstruction from point clouds and ground plans.
\newblock {\em International archives of photogrammetry remote sensing and
  spatial information sciences}, 34(3/W4):37--44, 2001.

\bibitem{vosselman2004recognising}
G.~Vosselman, B.~G. Gorte, G.~Sithole, and T.~Rabbani.
\newblock Recognising structure in laser scanner point clouds.
\newblock {\em International archives of photogrammetry, remote sensing and
  spatial information sciences}, 46(8):33--38, 2004.

\bibitem{Wang2017OCNN}
P.-S. Wang, Y.~Liu, Y.-X. Guo, C.-Y. Sun, and X.~Tong.
\newblock O-cnn: Octree-based convolutional neural networks for 3d shape
  analysis.
\newblock {\em ACM Transactions on Graphics (SIGGRAPH)}, 36(4), 2017.

\bibitem{wu20153d}
Z.~Wu, S.~Song, A.~Khosla, F.~Yu, L.~Zhang, X.~Tang, and J.~Xiao.
\newblock 3d shapenets: A deep representation for volumetric shapes.
\newblock In {\em Proc. IEEE Conf. on Computer Vision and Pattern Recognition
  (CVPR)}, pages 1912--1920, 2015.

\bibitem{yang2018foldingnet}
Y.~Yang, C.~Feng, Y.~Shen, and D.~Tian.
\newblock Foldingnet: Point cloud auto-encoder via deep grid deformation.
\newblock {\em Proc. IEEE Conf. on Computer Vision and Pattern Recognition
  (CVPR)}, 2018.

\bibitem{yi2016scalable}
L.~Yi, V.~G. Kim, D.~Ceylan, I.~Shen, M.~Yan, H.~Su, A.~Lu, Q.~Huang,
  A.~Sheffer, L.~Guibas, et~al.
\newblock A scalable active framework for region annotation in 3d shape
  collections.
\newblock {\em ACM Transactions on Graphics (TOG)}, 35(6):210, 2016.

\end{thebibliography}
}

\end{document}